\PassOptionsToPackage{capitalize}{cleveref}
\PassOptionsToPackage{section}{placeins}
\documentclass{applemlr}
\input{arxiv_preamble}

\usepackage{amsmath,amsfonts,bm}

\def\eqref#1{equation~\ref{#1}}

\def\1{\bm{1}}

\DeclareMathAlphabet{\mathsfit}{\encodingdefault}{\sfdefault}{m}{sl}
\SetMathAlphabet{\mathsfit}{bold}{\encodingdefault}{\sfdefault}{bx}{n}

\usepackage{titletoc}
\usepackage{url}

\usepackage{graphicx}
\usepackage{wrapfig}
\usepackage{tikz}
\usetikzlibrary{arrows.meta,shapes.geometric}
\definecolor{cCost}{HTML}{4A2A82}
\definecolor{cPrec}{HTML}{3A8C52}
\definecolor{cRank}{HTML}{E6A845}
\definecolor{cDepth}{HTML}{595959}
\definecolor{gGQAbf}{HTML}{7EC19B}
\definecolor{gGQAq}{HTML}{3A8C52}
\definecolor{gMLAbf}{HTML}{FED286}
\definecolor{gMLAq}{HTML}{E6A845}
\definecolor{gKIVI}{HTML}{5497C1}
\tikzset{gsolid/.style={line width=0.9pt}, gdash/.style={line width=0.9pt, dash pattern=on 1.8pt off 1.1pt}}
\newcommand{\gbox}[1]{\makebox[1.7em][l]{#1}}
\newcommand{\gsq}[2]{\gbox{\tikz[baseline=-0.55ex]{\draw[#1,#2] (0,0)--(0.45,0);
  \filldraw[draw=#1,fill=white,line width=0.6pt] (0.225,0) +(-0.06,-0.06) rectangle +(0.06,0.06);}}}
\newcommand{\gci}[2]{\gbox{\tikz[baseline=-0.55ex]{\draw[#1,#2] (0,0)--(0.45,0); \fill[#1] (0.225,0) circle (0.07);}}}
\newcommand{\gstar}[1]{\gbox{\tikz[baseline=-0.55ex]{\draw[#1,gsolid] (0,0)--(0.45,0);
  \node[star,star points=5,star point ratio=2.3,fill=#1,draw=black!80,line width=0.25pt,inner sep=0pt,minimum size=2.7mm] at (0.225,0) {};}}}
\newcommand{\gplus}[1]{\gbox{\tikz[baseline=-0.55ex]{\draw[#1,gsolid] (0,0)--(0.45,0);
  \draw[#1,line width=1.5pt] (0.155,0)--(0.295,0) (0.225,-0.07)--(0.225,0.07);}}}
\newcommand{\gnone}{\gbox{}}
\newcommand{\kvcube}{KV-Kaizen}
\newcommand{\kvcubeshort}{KVK}
\usepackage{booktabs}
\usepackage{multirow}
\usepackage{amsmath}
\usepackage{amssymb}

\crefname{appsec}{Appendix}{Appendices}
\Crefname{appsec}{Appendix}{Appendices}

\title{\kvcube: Learning Context-Adaptive \\Cache Compression Choices}

\metadata[SL$^\star$]{\sffamily Work done as an intern at Apple.}
\author{Jo\~ao Monteiro}
\author{Louis B\'ethune}
\author{Anastasiia Filippova}
\author[\star]{Sonia Laguna}
\author{David Grangier}
\author{Marco Cuturi}

\affiliation{Apple}

\abstract{As the context size of text processed with an LLM grows, the size of KV caches can outstrip the memory allocated for the original model weights. 
This impacts LLM throughput negatively, since decoding is memory-bound and decode cost grows with cache size.
Recent work alleviates this bottleneck by discarding the least relevant tokens.
\textit{Eviction} introduces a tension, since a one-off decision to discard content may prove detrimental later. 
Instead, we focus on alternative choices that can lead to cache compression \textbf{without} evicting tokens.
We achieve this by learning a selector that is able to produce, based on context, a per-layer cache configuration towards an overall compression budget.
The selector operates along three axes: sharing one cache across layers (\emph{depth}), caching at fewer bits (\emph{precision}), or truncating the low-rank latent cache representations (\emph{rank}). We call the resulting method \kvcube, for the many small per-layer choices it compounds.
We observe that these interventions taken independently and uniformly over all layers limit achievable compression because they degrade accuracy.
Crucially, composing them \textit{locally} and \textit{adaptively} to the context can instead preserve accuracy while achieving large memory savings.
At inference, the selector runs once, before pre-fill.
In evaluations on instruction following and reasoning tasks, our selectors reach the Pareto frontier of accuracy against cache size, against learning-free and \textit{post-hoc} baselines. 
On long-context tasks, \kvcube{} improves on eviction and can be composed with it, reaching a $32\times$ smaller decode-time cache on a $14$B model while preserving accuracy.
A $4\times$ cache size reduction incurs no accuracy degradation from $7$B parameters up, and a compressed model is more accurate than a smaller uncompressed one with the same cache size. Together, these findings support pre-training large models and compressing them only afterwards.\looseness-1
}
\metadata[Correspondence]{\sffamily
Jo\~ao Monteiro: \url{jmonteiro2@apple.com};
Louis B\'ethune: \url{l_bethune@apple.com};
Anastasiia Filippova: \url{a_filippova@apple.com};
Sonia Laguna$^\star$: \url{slaguna@ethz.ch};
David Grangier: \url{grangier@apple.com};
Marco Cuturi: \url{m_cuturi@apple.com}.
}

\begin{document}

\maketitle

\section{Introduction}\label{sec:intro}

Attention-based LLMs trade memory for compute: a KV cache stores the key and value (KV) vectors of every prefilled or generated token, so that they are not recomputed at each decoding step.
The memory this takes grows with the length and number of the sequences served.
\texttt{Qwen2.5-14B}~\citep{qwen25}, for instance, uses a cache memory of $192$\,KiB per token, so serving $32$ sequences of $8$k tokens fills up $48$\,GiB, more than the model weights in \texttt{bf16}.
Naturally, that memory cache introduces a bottleneck: every decoded token reads all of the KV cache, so decoding is limited by memory bandwidth rather than by compute.
A smaller cache therefore lets a model serve more and longer sequences, and decode at a higher throughput.
The challenge thus becomes caching more efficiently without incurring information loss with respect to the original cache.

\textbf{Smaller KVs.} To alleviate these issues, recent works shrink KV caches by keeping positions that are deemed relevant, and drop the rest~\citep{zhang2023h2o,cai2024pyramidkv,qincake}. When such \textit{eviction} strategies are query-independent, they run the risk of sacrificing tokens that could have proved useful for other queries. This failure can be mitigated by dropping KV at inference conditionally on a specific query~\citep{li2024snapkv}, or by guiding eviction using a wide-reaching generic prompt~\citep{zhu2025oraclekv}.
~\citet{eyuboglu2026cartridges,zweiger2026fast,monteiro2026nectar} have also proposed to learn \textit{compact} KV cache representations of a fixed context one attends to repeatedly, by distilling the LLM's use of KV caches against a wide set of synthesized QAs relevant to the targeted context.
A natural counterpart to these \textit{compaction/eviction} approaches is cache \textit{compression} without token selection. For instance, one can store the KV cache in low precision to reduce the amount of information that must be moved at every cache read~\citep{hooper2024kvquant,liu2024kivi}, or target lower dimensional representations of KV states, as proposed with multi-head latent attention (MLA)~\citep{deepseekv2}.
Since the KV cache encodes tokens at every head and every layer, cache redundancies have been exploited so that groups of layers share the same KV states rather than producing their own cache~\citep{brandon2405reducing,sun2024you,rcla2026}, as recently showcased in \texttt{Gemma-4}~\citep{team2026gemma} and \texttt{DeepSeek-V4.1-Flash}~\citep{deepseekai2026deepseekv41flash}. However, applying these structural changes aggressively quickly results in performance degradation.%

\textbf{Learned Adaptive KV Compression Strategies.}  In this work, we compose three \textit{compression} approaches: MLA's dimension or \emph{rank}, the cache \emph{precision}, and the effective \emph{depth} determined by how layers are grouped to share a cache. We refer to this method as \emph{\kvcube}, in reference to the ``kaizen'' management method that leverages minor, gradual adjustments to improve an existing pipeline. The savings of these three approaches compound, combining multiple mild compression that yield high compression ratios that would prove too lossy if carried by a single direction. 
We learn by how much to push along each of precision, rank, and depth during supervised fine-tuning: a small \emph{selector} reads the prompt and picks one option per layer. During inference, the selector runs once at the start of pre-fill, and its choice then holds for every generated token. Since we train selector and model together, the model adapts to the cache it will be served under. We evaluate \kvcube{} on a range of models covering $1.5$B to $32$B parameters and different model families, on instruction following, reasoning tasks and long-context retrieval and question answering, against learning-free, \textit{post-hoc} and manually designed baselines at matched cache sizes. Our contributions are:
\begin{figure}[t]
\centering
\includegraphics[width=\linewidth]{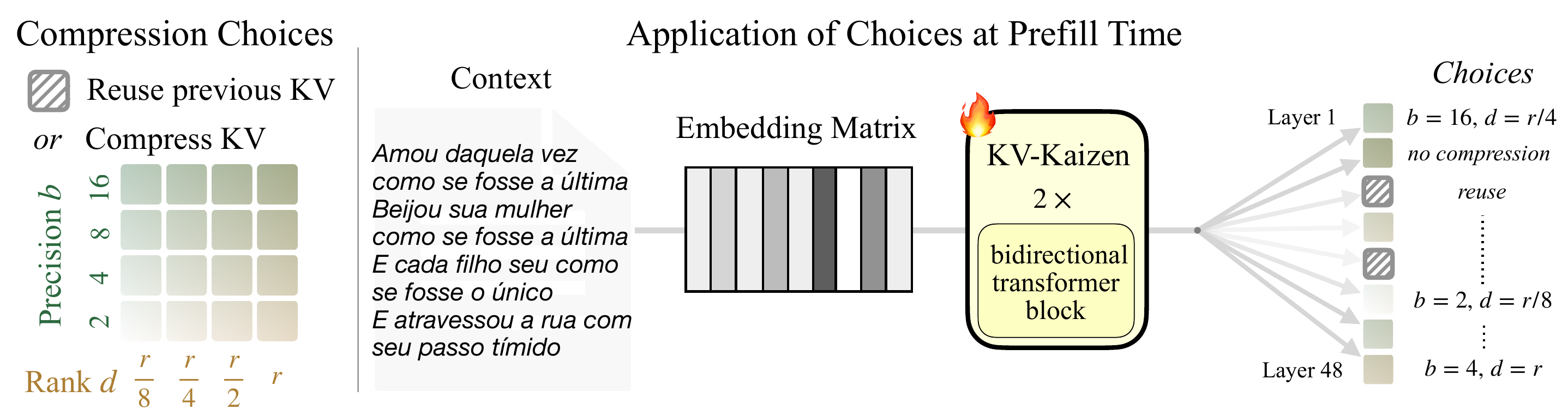}
\caption{\textbf{Per-layer cache selection and composition of compression axes.} The selector chooses among $16$ width--precision pairs or, alternatively, a inherit/reuse action per layer. Producing a cache costs $(d+d_R)b$ bits per token; inheritance costs zero. The configuration is fixed during generation, including at prefill time. In theory, the maximal compression factor that could be attained for a 48 layers model such as \texttt{Qwen2.5-14B} could be in practice $8\times$ (precision) times $10.7$ (lowest rank), and depth $48$, by reusing the first KV layer at all subsequent layers (\Cref{tab:ranges}). This would yield a maximal compression of $4108\times$, though all of our experiments target far lower factors.}
\vspace{-0.5cm}
\label{fig:method}
\end{figure}
\begin{list}{$\bullet$}{\leftmargin=1.3em \itemindent=0pt \labelwidth=0.8em \labelsep=0.5em
  \topsep=2pt \partopsep=0pt \parsep=0pt \itemsep=1pt}
\item We formulate depth, rank, and precision as a single per-layer choice under a shared budget, with all actions expressed in a common cost of bits per token. The budget can also be an input, so that one selector serves several (\Cref{sec:eval_budget_cond}).
\item \emph{Cache size reductions due to rank, precision, and depth compose.} We show that mild settings of the three cache reduction approaches we consider reach compression ratios none reaches alone. For instance, a $4\times$ cache size reduction gives accuracy comparable to the corresponding backbone in the evaluated models of at least $7$B parameters. Trained selectors outperform the static and random configurations as shown in \Cref{fig:headline} (\Cref{sec:eval_training,sec:eval_axes}).
\item \emph{Token eviction composes with the non-eviction approaches we consider.} As we do not evict tokens, eviction can be applied to the same cache and the two ratios multiply. On \texttt{Qwen2.5-14B} at $16$k, a $4\times$ compression due to our cache configuration plus eviction to one position in eight gives a $32\times$ smaller cache at decode time, with no significant accuracy difference (\Cref{sec:eval_longctx}).
\item \emph{Cache size reductions realize gains on a real device.} Benchmarking the three approaches on device, we observe that composing them keeps peak memory nearly flat as sequences grow, and that most combinations decode faster than an uncompressed cache (\Cref{sec:bmk}).
\end{list}

\section{Composing compression axes under one budget}\label{sec:method}

\subsection{Background and notation}\label{sec:background}

We consider a model of $L$ layers and denote by $l$ the layer index. Under grouped-query attention (GQA)~\citep{ainslie2023gqa}, layer $l$ stores $K_l$ and $V_l$ for $n_{\text{kv}}$ key-value heads of dimension $d_h$, so its cache holds $2 n_{\text{kv}} d_h$ scalars per token. We consider three cache compression approaches, none of which drops input tokens. \textbf{Depth.} Under cross-layer attention~\citep{brandon2405reducing,rcla2026}, a layer uses its own query but reads another layer's keys and values, where depth is the number of layers producing their own KV cache. In other words, each layer either produces its own cache, in which case we call it an \emph{anchor}, or inherits from the nearest anchor before it. \textbf{Rank.} Multi-head latent attention (MLA)~\citep{deepseekv2} stores one latent representation $c^{KV}_l \in \mathbb{R}^{r}$ shared across heads, and a decoupled key $k^{R}_l \in \mathbb{R}^{d_R}$ reserved for RoPE~\citep{su2024roformer}, carrying positional information. A layer holds $r + d_R$ values per token instead of $2 n_{\text{kv}} d_h$. Given our focus on methods that enable cache reduction after pre-training, we convert non-MLA models into MLA ones using CARE~\citep{care2026}. The \emph{rank} $r$ is the dimension of the MLA latent representation of the KV states, and we can control the cache size by truncating $c^{KV}_l$ to its first $d \le r$ coordinates, giving an effective latent rank $d$. This truncation approach works well to the extent that the leading coordinates carry the most information. Training nested representations encourages this ordering~\citep{kusupati2022matryoshka}, but a converted model does not necessarily do so (\Cref{app:worked}). This can be accounted for during fine-tuning. \textbf{Precision.} A cached tensor can be held at $b\leq16$ bits, by symmetric uniform quantization with one scale per token (\Cref{app:experimental_details}). During fine-tuning, the cache is quantized and dequantized in the forward pass with the gradient passed straight through, so the model trains against and learns to cope with a $b$-bit representation of KV states.

\subsection{The selector action space and the cache configuration cost}\label{sec:action_space}

We treat depth, rank, and precision as a single per-layer choice (\Cref{fig:method}), since all three reduce to one quantity: the number of bits a layer writes per token. The selector influences the cache cost via the following \emph{actions}: \emph{inherit} the cache, in which case the layer will simply re-use the most recently produced cache, incurring a cost of $0$. If instead a layer produces a cache, then actions correspond to deciding a pair $(d,b)$ of a kept width and a bit-width, with cost
\begin{equation}\label{eq:cost}
  c(a) =
  \begin{cases}
    0 & a = \text{inherit},\\
    (d + d_R)\cdot b & a = (d, b),
  \end{cases}
\end{equation}
in bits per token, where $d$ and $d_R$ are MLA's kept latent width and positional key width, as discussed in \Cref{sec:background}. As a non-MLA backbone has no latent to truncate and no separate key channel, the cost in that case follows with $d = 2 n_{\text{kv}} d_h$ and $d_R = 0$, the rank axis is unavailable, and a layer chooses only its bit-width and whether it produces or inherits a cache.

A \emph{plan} $a_{1:L}$, one action per layer, then has one scalar cost, which we report as a compression ratio against the unmodified \textsc{bf16} non-MLA cache,
\begin{equation}\label{eq:compression}
  \rho(a_{1:L}) = \frac{C_0}{\sum_{l=1}^{L} c(a_l)},
  \qquad
  C_0 = L \cdot 2\, n_{\text{kv}} d_h \cdot 16,
\end{equation}
with $n_{\text{kv}}$ KV heads of dimension $d_h$. Layer $1$ must own a cache, so no plan can cost nothing. A GQA layer at $b=16$ costs exactly $C_0/L$. An MLA layer at full width already costs less, since it represents the cache in smaller dimensions, and so $\rho$ in that case starts between $1.60\times$ and $1.88\times$ depending on the underlying model before the selector chooses anything, as detailed in \Cref{tab:ranges}. That part of the compression comes from the conversion and not from the cache configuration and, as such, one must be extra careful when comparing MLA vs. non-MLA cache configurations.

If $\mathcal{B}$ is the bit-widths the precision axis can aim for, $\{2,4,8,16\}$, and $\mathcal{D}$ the kept widths the rank axis can reach, i.e., the model's own $r$ successively halved, $\{r/8, r/4, r/2, r\}$, then the action space is
\begin{equation}\label{eq:action_space}
  \mathcal{A} \;=\; \{\text{inherit}\} \;\cup\; \{\, (d,b) \;:\; d \in \mathcal{D},\ b \in \mathcal{B} \,\},
  \qquad
  |\mathcal{A}| \;=\; |\mathcal{D}| \cdot |\mathcal{B}| + 1 .
\end{equation}
That gives the selector $17$ actions per layer as illustrated in \Cref{fig:method}. Restricting the action space to a subset of these three approaches recovers the corresponding single-axis methods since \emph{precision} is per-layer cache quantization, \emph{rank} is per-layer rank truncation, and \emph{depth} is learned cross-layer sharing. Selectors operating on pairs of these approaches can also be defined by simply setting $\mathcal{B}=\{16\}$ or $\mathcal{D}=\{r\}$. \Cref{tab:ranges} shows the cache reduction factors each such subset of $\mathcal{A}$ can reach.

\subsection{Learning the cache configuration against a budget}\label{sec:selector}

\begin{wrapfigure}{r}{0.5\textwidth}
\vspace{-\baselineskip}
\centering
\includegraphics{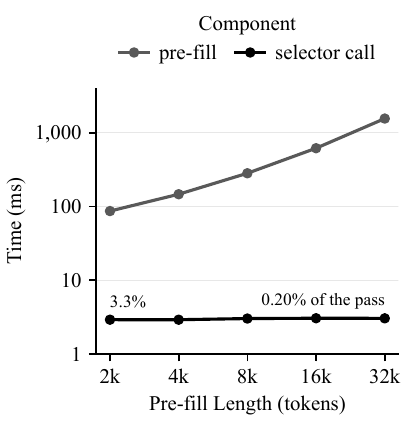}
\caption{\textbf{Selector and pre-fill latency versus input length.} Measurements use \texttt{Qwen2.5-7B} at $4\times$ compression, batch size $1$, on one H100. The selector takes about $3$\,ms, or $3.3\%$--$0.20\%$ of pre-fill latency.}
\label{fig:prefill}
\end{wrapfigure}
A selector $s:\mathcal{V}^T \to \mathcal{A}^L$ maps $T$ input tokens from the vocabulary $\mathcal{V}$ into per-layer cache configurations, and learns to do so while the backbone trains. We parameterize the selector with one linear head per layer, each mapping the shared feature to logits over the action space~$\mathcal{A}$. When depth is not within the action space of the selector being trained, its logit is forced to $-\infty$, which includes layer $1$. Selection uses Gumbel-softmax~\citep{jang2016categorical,maddison2016concrete} with a straight-through estimator~\citep{bengio2013estimating}, hard $\arg\max$ forward and softmax backward at a temperature annealed from $2.0$ to $0.1$, and a plain $\arg\max$ at evaluation. The heads are initialized so that every layer's $\arg\max$ is the least-compressed action, so training starts from the uncompressed model and departs from it as training progresses.

The shared feature comes from a small model that takes the prompt as an input. It is made of two bidirectional transformer blocks that ingest the backbone's own token embeddings. The representations are later averaged along the time axis. The evaluated selectors range from $2.6$ to $3.7$M parameters, which represents less than $0.05\%$ of the model from $7$B upward. This sequence model shared across layers runs once at pre-fill, as depicted in \Cref{fig:prefill} where the selector takes about $3$\,ms, from $2$k to $32$k input tokens, representing a share ranging from $3.3\%$ to $0.20\%$ in time to first token, while the cache reduction applies at every memory-bound decode step. The plan is then held for the whole generation, so the cache shape is stable across decode steps and the selector costs nothing per decoded token. In our implementation, training with the selector takes $1.0$ to $3.1\times$ as long as a plain fine-tune (\Cref{app:selector_cost}). The selector is trained jointly with the backbone under the language modeling loss plus a term pushing realized cost toward a target compression $\rho_\star$:
\begin{equation}\label{eq:rate}
  \mathcal{L}_{\text{rate}} = \beta \cdot \mathbb{E}_i \left| \frac{1}{\rho_i} - \frac{1}{\rho_\star} \right|,
  \qquad \frac{1}{\rho_i} = \frac{1}{C_0}\sum_{l} \sum_{a \in \mathcal{A}} w^{(i)}_{l,a}\, c(a),
\end{equation}
\[
  \mathcal{A} \;=\; {\color{cDepth}\underbrace{\color{black}\{\text{reuse}\}}_{\text{depth}}}
  \;\cup\; \Big( {\color{cRank}\underbrace{\color{black}\{\tfrac18,\tfrac14,\tfrac12,1\}}_{\text{rank (fraction kept)}}}
  \times {\color{cPrec}\underbrace{\color{black}\{2,4,8,16\}}_{\text{precision (bits)}}} \Big)
\]
for input sequence $i$, where $w^{(i)}_{l}$ is layer $l$'s straight-through one-hot selection vector and $c(a)$ is a lookup cost, so that $1/\rho_i$ evaluates the total cost of the plan cached. We introduce a hyperparameter $\beta$ that must be tuned, and its best value depends both on the backbone and $\rho_\star$. Sweeping it for every configuration is expensive, so we adapt it during training instead. We raise $\beta$ while the requested compression is missed and relax it once it is met. This works well enough that one setting reaches every fixed target we train for, which is what lets us compare a grid of configurations. We discuss the $\beta$ adaptive rule in full in \Cref{app:budget_controller}, and compares it against a swept constant.

In our evaluations discussed in \Cref{sec:eval}, we compare learned selectors with static cache configurations \emph{at a matched budget and over the same action space} (\Cref{app:experimental_details}). Similarly, we consider picking the cache configuration at random out of a set of options that yield the desired target cache size. In all cases, models are fine-tuned with the cache configuration already in place, under exactly the recipe of the learned configurations, since applying a configuration to a model that never trained under it degrades it systematically (\Cref{tab:posthoc}).

\section{Evaluation}\label{sec:eval}

\subsection{Setup}\label{sec:eval_setup}

We reduce the cache size of already trained models by fine-tuning them along with the \kvcube{} selector. The base models we start from are listed in \Cref{tab:backbones} in \Cref{app:experimental_details} along their respective attention patterns. For configurations that use the rank axis we rely on MLA backbones, converted from the original model with CARE~\citep{care2026}. For all models we use the same SFT recipe: the Dolci instruction mixture~\citep{dolci} at length $4096$ for $270{,}000$ steps, with the selector and the backbone trained together. \Cref{app:experimental_details} gives the optimizer settings, the conversion settings and the selector architecture in full. We report accuracy on an instruction following task and a reasoning task, IFEval~\citep{zhou2023instruction} and GSM8K~\citep{cobbe2021training}, both through the lm-evaluation-harness~\citep{eval-harness}, and we evaluate longer contexts on RULER~\citep{hsieh2024ruler} at $16$k. We implement all baselines and compare realized cache reduction factors, matching cache sizes where available and stating both factors otherwise. The cost model in \cref{eq:cost,eq:compression} excludes quantization metadata; \Cref{app:action_space} reports its storage overhead.

\begin{figure}[t]
\begin{center}
\includegraphics[width=0.9\textwidth]{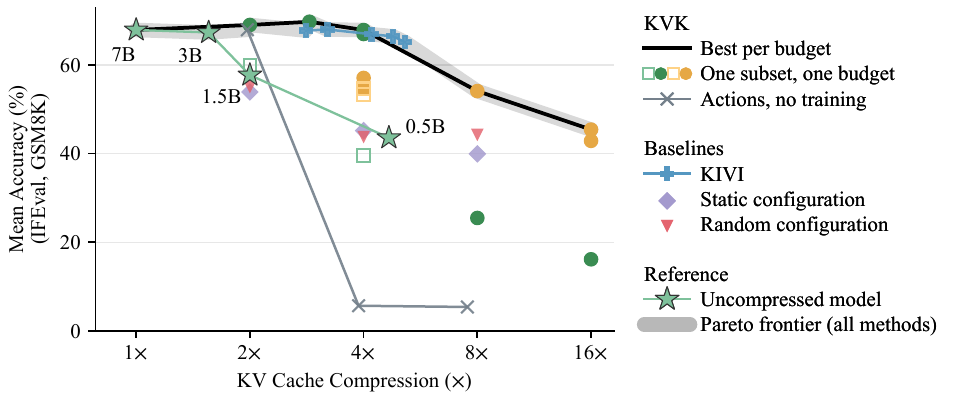}
\end{center}
\caption{\textbf{Accuracy versus cache compression on \texttt{Qwen2.5-7B}.} Accuracy is the mean of IFEval and GSM8K; compression is relative to the full $7$B \textsc{bf16} cache. The starred curve shows uncompressed models at four sizes. The black curve is the best \kvcubeshort{} run per budget; the pale band gives the Pareto frontier of all methods shown. Static and random configurations are fine-tuned. KIVI: grouped, \textsc{bf16} residual.}
\label{fig:headline}
\end{figure}

\subsection{Reducing the cache size by more than about $5\times$ needs training}\label{sec:eval_training}

\Cref{fig:headline} compares mean IFEval and GSM8K accuracy against cache compression on \texttt{Qwen2.5-7B}. For example, a factor of $2\times$ corresponds to half the cache size of the uncompressed model. Remarkably, at a cache reduction factor of $4\times$, the \kvcubeshort{} selector reaches the same accuracy as the uncompressed control. \textbf{Simple baselines}. The static configuration (the same action applied uniformly to every layer) and the random configuration (one action drawn at random per layer) both degrade accuracy, and fall below the frontier in \Cref{fig:headline} at the same cache size. In \Cref{tab:forced-depth} we separately compare \textit{learned}, \textit{evenly spaced}, and \textit{random anchor} placement at a fixed cache size. In \Cref{app:regret_diagnostic} we compare the learned configuration with $24$ alternatives at similar cache cost, while holding model weights fixed. \textbf{Search baseline}. A cache configuration can also be found by search rather than learned. In \Cref{app:regret_diagnostic} we compare an offline dynamic program over the same action space against the uniform and random configurations, on held-out calibration data at a similar cache reduction factor of $8\times$. The program's configuration raises the language modeling loss by $0.030$, while the uniform configuration raises it by $4.00$ and the best out of $64$ random configurations raises it by $1.25$. At fixed cache size, the loss depends on the configuration chosen.

The model must be train under the configuration that will be used for serving at inference time. Applying the configuration of a learned selector to a frozen backbone collapses accuracy: we see those measurements sitting at the bottom of \Cref{fig:headline}. \textit{Post-hoc} quantization can work with a more accurate quantizer: a KIVI-style scheme~\citep{liu2024kivi} applied after training stays on the frontier and reaches a cache reduction factor of about $5\times$, with a limit of two bits per stored entry. Beyond that compression factor of $5\times$, all configurations on the frontier are the learned ones. We report the individual numbers in \Cref{tab:posthoc} in \Cref{app:full_results}.  

We compare compression by serving smaller models (\Cref{fig:headline}). The starred curve shows \texttt{Qwen2.5} models at $0.5$B, $1.5$B, $3$B and $7$B, reporting their \textit{uncompressed} KV-cache (of lower dimension) against the $7$B reference. Past a factor of $2\times$ these models score below the Pareto front, making a compressed KV-cache of a $7$B model more accurate than the full KV-cache of smaller alternatives.  

\subsection{Scaling the base model size}\label{sec:eval_axes}

\begin{table}[t]
\centering\small\setlength{\tabcolsep}{5pt}
\caption{\textbf{Mean IFEval and GSM8K accuracy differences at $4\times$ cache reduction.} The $3$B, $7$B, and $14$B columns use \texttt{Qwen2.5}; Ref. identifies the size-matched control (MLA for rank, non-MLA otherwise). The first row gives absolute accuracies; subsequent rows give differences from Ref., with higher values better. Bold marks differences $\geq-0.053$, ie the largest observed seed range (\Cref{tab:seeds}).}
\label{tab:axes}
\begin{tabular}{llccccc}
\toprule
& & \multicolumn{5}{c}{Mean accuracy / difference $\uparrow$} \\
\cmidrule(lr){3-7}
& & \multicolumn{3}{c}{\emph{Qwen2.5}} & \multicolumn{2}{c}{\emph{Other families}} \\
\cmidrule(lr){3-5}\cmidrule(l){6-7}
Approaches & Ref. & 3B & 7B & 14B & Mistral-7B & OLMo-3-7B \\
\midrule
\gstar{gGQAbf}Uncompressed control & -- & 0.673 & 0.679 & 0.723 & 0.349 & 0.508 \\
\midrule
\gci{gGQAq}{gsolid}Precision & Non-MLA & $\mathbf{-0.041}$ & $\mathbf{-0.009}$ & $\mathbf{+0.040}$ & $\mathbf{+0.079}$ & $\mathbf{+0.011}$ \\
\gsq{gGQAbf}{gdash}Depth & Non-MLA & $-0.354$ & $-0.283$ & $-0.147$ & $-0.092$ & $-0.126$ \\
\gci{gGQAq}{gdash}Precision + depth & Non-MLA & $-0.063$ & $\mathbf{-0.000}$ & $\mathbf{+0.031}$ & $\mathbf{+0.068}$ & $\mathbf{+0.015}$ \\
\midrule
\gstar{gMLAbf}MLA conversion only & Non-MLA & $-0.178$ & $-0.105$ & $-0.091$ & $\mathbf{+0.005}$ & $\mathbf{-0.040}$ \\
\gsq{gMLAbf}{gsolid}Rank & MLA & $-0.219$ & $\mathbf{-0.028}$ & $\mathbf{-0.005}$ & $\mathbf{-0.007}$ & $\mathbf{+0.006}$ \\
\gci{gMLAq}{gsolid}Precision + rank & MLA & $-0.167$ & $\mathbf{-0.003}$ & $\mathbf{+0.043}$ & $\mathbf{-0.001}$ & $\mathbf{+0.009}$ \\
\gsq{gMLAbf}{gdash}Depth + rank & MLA & $-0.224$ & $\mathbf{-0.041}$ & $\mathbf{-0.008}$ & $\mathbf{-0.033}$ & $\mathbf{+0.020}$ \\
\gci{gMLAq}{gdash}All three & MLA & $-0.169$ & $\mathbf{-0.028}$ & $\mathbf{+0.021}$ & $\mathbf{+0.027}$ & $\mathbf{+0.000}$ \\
\bottomrule
\end{tabular}
\end{table}

We fix the base model at $7$B in \Cref{fig:headline}. We sweep the size of the base model, and the set of cache reduction approaches the selector is allowed to use. We also add two other model families, \texttt{Mistral-7B}~\citep{jiang2023mistral7b} and \texttt{OLMo-3-7B}~\citep{dolci}, so that the results do not rest on \texttt{Qwen2.5} alone. In \Cref{tab:axes} we look at the accuracy gap against the uncompressed control in the same attention class (higher is better $\uparrow$). That is, MLA models are compared against an MLA control fine-tuned with the same recipe and no compression, and non-MLA models against a non-MLA control. The gap closes as the base model grows. At $3$B every configuration sits behind its control, and at $7$B and $14$B depth on its own is the only one that is not level with its control or ahead of it. \Cref{tab:scaling} in \Cref{app:seeds_scaling} carries the same comparison out to $32$B, and also reads it against the model we start from rather than against our own control, which gives the same picture.

\begin{figure}[t]
\begin{center}
\includegraphics[width=0.85\textwidth]{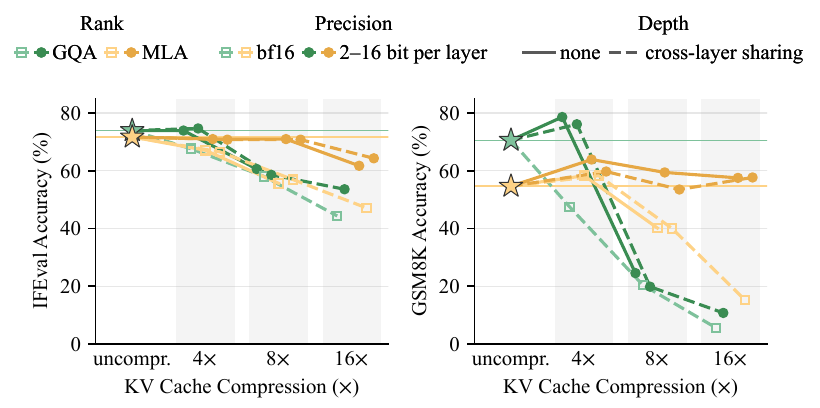}
\end{center}
\caption{\textbf{IFEval and GSM8K accuracy versus realized cache compression on \texttt{Qwen2.5-14B}.} Each curve uses a different subset of compression axes: hue marks rank, shade and marker mark precision, and a dashed line marks depth. Horizontal lines mark the GQA and MLA controls.}
\label{fig:frontier-14b}
\end{figure}

We also note that combining approaches does not add their costs together. On the non-MLA backbone, precision on its own and precision combined with depth reach similar gaps at every size, while depth on its own, which has to let three layers in four inherit to reach $4\times$, is far behind both. At $14$B this is a comparison at one cache size: with only depth available the selector loses accuracy at a factor of $4\times$, where precision and depth together at that same cache size land above the control. Allowing several approaches lets the selector meet the budget without relying on aggressive depth compression. \Cref{tab:axes-7b} in \Cref{app:full_results} breaks the same configurations down per task at $7$B, and depth is observed to cost more on reasoning than on instruction following. \Cref{fig:frontier-14b} reports the same configurations at $14$B, per task. Beyond $8\times$, selector settings unable to act on rank lose reasoning accuracy first. Depth on its own at $16\times$ still reaches $0.444$ on instruction following, but only $0.056$ on reasoning.

\subsection{Long context evaluations and composition with eviction methods}\label{sec:eval_longctx}

\begin{table}[t]
\centering\small\setlength{\tabcolsep}{5pt}
\caption{\textbf{RULER accuracy at $\bm{16}$k on \texttt{Qwen2.5-14B}.} NIAH multi-key and QA report retrieval and question-answering accuracy over $275$ documents each. $\rho$ is the decode-time cache reduction relative to the uncompressed GQA model; keep is the retained token fraction, and $G$ the quantization group size. Parentheses give QA differences from the SFT control. Single-needle accuracy is $1.000$ throughout and therefore omitted from the table.}
\label{tab:longctx}
\begin{tabular}{lrcc}
\toprule
& & \multicolumn{2}{c}{Accuracy $\uparrow$} \\
\cmidrule(lr){3-4}
Configuration & $\rho\uparrow$ & NIAH multi-key & QA (SQuAD) \\
\midrule
\gnone Untuned, before our recipe & $1.0$ & 1.000 & 0.592\,($+0.049$) \\
\gstar{gGQAbf}Uncompressed control & $1.0$ & 0.956 & 0.543 \\
\midrule
\gci{gGQAq}{gdash}\textbf{\kvcubeshort}, $4\times$ (precision + depth) & $4.0$ & 0.967 & 0.614\,($+0.071$) \\
\gnone\quad \textbf{+ SnapKV}, keep $0.25$ & $16.0$ & 0.967 & 0.560\,($+0.017$) \\
\gnone\quad \textbf{+ SnapKV}, keep $0.125$ & $32.0$ & 0.967 & 0.531\,($-0.012$) \\
\midrule
\gplus{gKIVI}KIVI 4-bit $G$=128 & $3.7$ & 0.971 & 0.515\,($-0.028$) \\
\gnone\quad + SnapKV, keep $0.125$ & $29.5$ & 0.949 & 0.464\,($-0.079$) \\
\midrule
\gnone SnapKV, keep $0.25$ & $4.0$ & 0.935 & 0.501\,($-0.042$) \\
\gnone SnapKV, keep $0.125$ & $8.0$ & 0.916 & 0.478\,($-0.064$) \\
\gnone PyramidKV, keep $0.25$ & $4.0$ & 0.949 & 0.485\,($-0.057$) \\
\bottomrule
\end{tabular}
\end{table}

In \Cref{tab:longctx} we reports RULER at $16$k on \texttt{Qwen2.5-14B}. We use this setting for two things: to compare the approaches we consider against token eviction methods such as SnapKV~\citep{li2024snapkv} and PyramidKV~\citep{cai2024pyramidkv}, and to combine the two. The retrieval tasks can not discriminate the configurations, since every method solve single-needle (100\%) while the multi-key scores remain similar: therefore we focus on QA. At a cache reduction factor of $4\times$, our selector reaches $0.614$ on question answering, against $0.501$ for SnapKV (keeping one position in four) and $0.485$ for PyramidKV (at the same cache size). \Cref{tab:longctx-7b} in \Cref{app:full_results} shows similar results for a $7$B model.

As we do not drop tokens, an evictor can run on the same cache and the cache reduction factors due to each approach multiply. SnapKV at one position in four combined with our $4\times$ configuration gives a cache reduction factor of $16\times$ at an accuracy of $0.560$, and at one position in eight the factor reaches $32\times$ at $0.531$. Combining eviction with a \textit{post-hoc} quantizer instead is worse: four-bit KIVI under the same evictor holds a larger cache, $29.5\times$ against our $32\times$, and reaches a lower accuracy of $0.464$.

\subsection{One selector covers a set of budgets}\label{sec:eval_budget_cond}

\begin{table}[tb]
\centering\small\setlength{\tabcolsep}{4pt}
\caption{\textbf{Mean IFEval and GSM8K accuracy with budget-conditioned selectors on \texttt{Qwen2.5-7B}.} $\rho_\star$ is the requested cache reduction factor. Against $\{1,2,4\}$, the uncompressed control scores $0.679$ at $1\times$, and single-target selectors $0.690$ at $2\times$ (precision only) and $0.679$ at $4\times$. With all three approaches the MLA control scores $0.574$; requests are met within $2\%$ except $1.94\times$, $1.78\times$ and $1.90\times$ at $2\times$, in row order, and $3.81\times$ for the $540$k selector at $4\times$.}
\label{tab:budget-sets}
\begin{tabular}{lllcccc}
\toprule
& & & \multicolumn{4}{c}{Mean accuracy $\uparrow$ at requested $\rho_\star$} \\
\cmidrule(lr){4-7}
Approaches & Training & Steps & $1\times$ & $2\times$ & $4\times$ & $8\times$ \\
\midrule
\gci{gGQAq}{gdash}Precision + depth & $\{1,2,4\}$     & $270$k      & $\mathbf{0.703}$ & $\mathbf{0.659}$ & $\mathbf{0.612}$ & -- \\
\midrule
\gci{gMLAq}{gdash}All three         & One per target  & $270$k each & -- & $0.590$ & $0.547$ & $0.541$ \\
\gci{gMLAq}{gdash}All three         & $[2,8]$         & $270$k      & -- & $0.526$ & $0.487$ & $0.475$ \\
\gci{gMLAq}{gdash}All three         & $[2,8]$         & $540$k      & -- & $\mathbf{0.553}$ & $\mathbf{0.524}$ & $\mathbf{0.502}$ \\
\bottomrule
\end{tabular}
\end{table}

Experiments so far train a selector to reach a single target compression factor $\rho_\star$. To let one selector serve more than one cache size, we make the target an input to the selector, through a zero-initialized FiLM layer~\citep{perez2018film} on the pooled feature, and training is so that each input sequence draws its own target. Trained on $\{1,2,4\}$ with precision and depth, one selector covers all three, ahead of the uncompressed control at $1\times$ and $0.067$ behind a single-target selector at $4\times$ (\Cref{tab:budget-sets}).  
  
\textbf{A wider compression range requires more training}. In \Cref{tab:budget-sets} we compare the selectors trained over $[2,8]$ with all three approaches, against one selector per target. Trained for twice as long, a single selector trails them by $0.038$, $0.023$ and $0.039$ at $2\times$, $4\times$ and $8\times$, for only two thirds of the steps the three single-target selectors would take. Wider sets of targets do worse: they stop following the request and prevent the rate multiplier from settling (\Cref{app:budget_sets}).

\subsection{Cache configuration diversity}\label{sec:eval_prompt_dependence}

We benchmark the per-layer choices of selectors trained at $3$B, $7$B, and $14$B and at several target compression factors, on sample prompts from IFEval, GSM8K, and RULER of up to $16$k tokens (\Cref{app:replay}). Selectors adapt the configuration to the context, but some configurations are shared across prompts, so a selector returns fewer distinct configurations than the prompts it sees. Configurations also change from one dataset to another, and a cache configuration found on one dataset need not suit another. We observed that some selectors even return a global configuration that realizes $\rho_\star$ on every prompt, so they could be dropped at test time. Whether a selector can be replaced by a single configuration therefore depends on the selector and has to be checked, ideally on representative test data. The selector only costs about $3$\,ms (\Cref{fig:prefill}), negligible compared to pre-fill.    

We also note that rank, precision, and depth are not interchangeable. Precision and rank change a layer by degrees, a coarser quantizer holding the same values less exactly and a narrower latent keeping the leading coordinates of a wider one, so the layer goes on computing what it computed before. Depth has no setting in between: a layer that gives up its cache works from keys and values it did not produce. That is the larger change, and the selector uses it least as a consequence. Depth does however enable higher compression factors. Precision cannot pass $8\times$ (its two-bit floor), and rank knob is unavailable on non-MLA models, so every configuration we train past $8\times$ on a non-MLA backbone must give up at least half of its layers, and none could reach that factor otherwise.

\section{Benchmarking cache compression approaches}\label{sec:bmk}

\begin{figure}[!htb]
\begin{center}
\includegraphics[width=\textwidth]{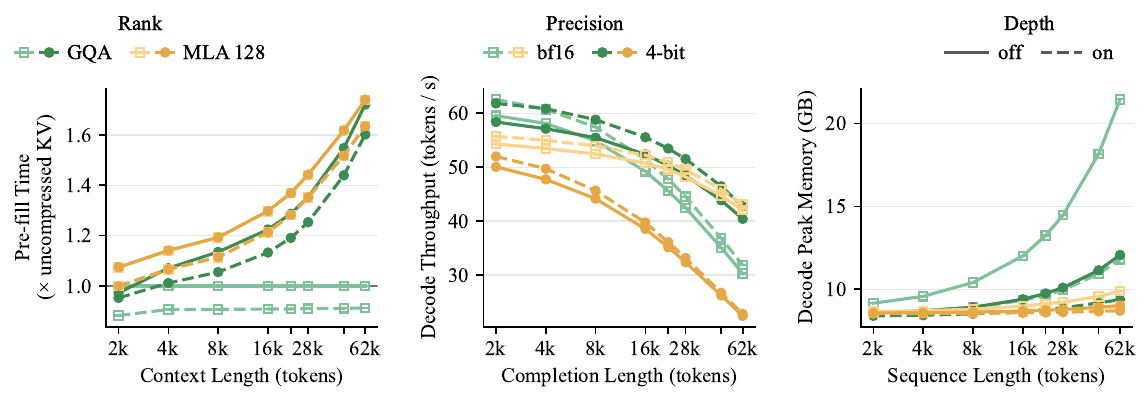}
\end{center}
\caption{\textbf{Pre-fill time, decode throughput and peak decode memory on an Apple M3 Ultra.} \texttt{Qwen2.5-14B} with $4$-bit weights at batch size $1$. Hue marks GQA or an MLA latent of rank $128$ (rank), shade and marker a \textsc{bf16} or $4$-bit cache (precision), and a dashed line $12$ groups of $4$ layers sharing a cache (depth). Pre-fill time is reported relative to the uncompressed cache in bf16 precision.}
\label{fig:ondevice}
\end{figure}

We benchmark memory and speed on Apple M3 Ultra with 512GB of RAM. We run \texttt{Qwen2.5-14B} with $4$-bit weights at batch size $1$, with the same approach on every layer: a $4$-bit cache ($3.56\times$), an MLA latent of rank $128$ ($10.7\times$), and $12$ groups of $4$ layers sharing one cache ($4\times$): a total of $152\times$ for all three. The $4$-bit cache uses the grouped quantizer of MLX~\citep{mlx2023}.

We study two synthetic scenarios: we sweep over context lengths to measure the impact on pre-filling (lower is better $\downarrow$) and we sweep over completions lengths measure decoding throughput (higher is better $\uparrow$). A \textsc{bf16} cache adds $12.3$\,GB of peak memory and the composed one $0.17$\,GB (\Cref{fig:ondevice}, right). The \textsc{bf16} baseline, which is the most expensive in memory, is also the fastest at pre-filling, a compute-intensive operation. The \textit{depth} knob always help in throughput. \textit{MLA} mostly pays off at long sequences lengths. The 4-bit KV-cache is more compute-intensive due to de-quantization cost, which increases the cost of pre-filling and MLA, but pays off at long horizons.

\section{Related work}\label{sec:related_work}
Most recent work focuses on dropping or merging token positions by attention score or recency~\citep{xiao2024efficient,li2024snapkv,zhang2023h2o,cai2024pyramidkv,qincake,shen122025lava,ge2023model}. An orthogonal set of approaches makes each position it keeps cheaper, and this is where the three cache reduction strategies we use come from: sharing KV states within a layer~\citep{shazeer2019fast,ainslie2023gqa} or between layers~\citep{brandon2405reducing,liu2024minicache,yang2024kvsharer,wang2025commonkv,mu2024crosslayer,wu2024layer,monteiro2024xc,wu2025systematic,rcla2026}, replacing them with a low-rank latent~\citep{deepseekv2,deepseekv3,transmla2025,care2026}, or holding each entry at fewer bits~\citep{hooper2024kvquant,liu2024kivi}. In most of these non-evicting approaches, the cache design is fixed at pre-training. Non-uniform allocation across layers is not new~\citep{cai2024pyramidkv,qincake,flexrank2026,adakv2026,squeezeattention2025,palu2024}, nor is a compression target in the objective~\citep{dmc2024}, and closest to us \citet{rcla2026} randomize the source layer during fine-tuning so one checkpoint tolerates sharing patterns chosen at test time. Prior work has looked into learned allocation of non-evicting cache compression. For instance, \citet{starkv2026} choose a rank per head and per block with a differentiable threshold and quantize the result, and \citet{matryoshkakv2025} train nested projections and then pick a rate per layer and head for a budget set after training. Both allocate within a narrower action space than ours, so a given cache reduction factor has to be reached by pushing fewer mechanisms further, where we claim a wider space lets the selector combine mild settings that cost less accuracy. A detailed account of relevant literature is reported in \Cref{app:related_work}.

\section{Conclusions, limitations, and future work}\label{sec:limitations}

We introduced \kvcube, which learns per-layer \textit{depth}, \textit{rank}, and \textit{precision} choices under one cache budget during SFT. Combining these approaches enables significant KV cache size reductions at little to no accuracy degradation. Moreover, at $16$k context length, combining $4\times$ representation compression with token eviction gives a $32\times$ smaller decode-time cache, with significant accuracy difference from the uncompressed control for a $14$B parameters model. We observed larger models to generally tolerate compression better, favoring the training of larger models with compression prior to deployment. We also noted that a budget-conditioned selector can serve several compression targets, although at a larger training cost. Some selectors vary with the input and between datasets, while others return the same configuration across prompts.

Fully realizing the memory and speed benefits of learned configurations, which mix widths and bit-widths across layers, requires specialized kernels. Rank compression also needs MLA conversion, whose accuracy cost varies across the evaluated models. Budget conditioning depends on the training targets: widening their range can reduce accuracy even at the less compressed settings. Finer control of the budget could let one model adapt to each deployment's hardware and accuracy needs.

\bibliography{bib}
\bibliographystyle{unsrtnat}

\newpage
\appendix
\crefalias{section}{appsec}
\crefalias{subsection}{appsec}

\section*{Appendix Contents}
\startcontents[appendix]
\printcontents[appendix]{}{1}{\setcounter{tocdepth}{2}}
\newpage
\section{Extended related work}\label{app:related_work}

\paragraph{Temporal eviction and selection.} Most recent work drops or merges tokens along the time axis. StreamingLLM~\citep{xiao2024efficient} keeps a sliding window plus attention sinks; SnapKV~\citep{li2024snapkv} uses attention scores to retain the chunks a prompt attends to; PyramidKV~\citep{cai2024pyramidkv}, CAKE~\citep{qincake} and LAVa~\citep{shen122025lava} vary the token budget across layers and heads; H\textsubscript{2}O~\citep{zhang2023h2o} and FastGen~\citep{ge2023model} manage cache updates during generation. The recurring difficulty is that token importance is query-dependent~\citep{zhu2025oraclekv}, so a position discarded for one query may be needed by the next. Recent methods mitigate this by predicting future relevance~\citep{zhu2025oraclekv}, re-scoring cheaply~\citep{hooper2025multipole} or reconstructing on demand~\citep{kim2025kvzip}. SALS~\citep{sals2025} composes the two directions rather than choosing between them, projecting the cache into a low-rank latent space and then selecting tokens inside that space with RoPE-free query-key products, so only the selected tokens are reconstructed. Attention may also be a weaker relevance signal than it looks: \citet{molfese2026exploring} find that ranking documents by cumulative attention score does not track task accuracy across their fine-tuned checkpoints. Our method leaves the time axis untouched, allowing it to be combined with token eviction. \citet{kvp2026} make the same point from the eviction side, naming representation compression as orthogonal and complementary to its learned eviction policy.

\paragraph{Architectural cache reduction and quantization.} Multi-query~\citep{shazeer2019fast} and GQA~\citep{ainslie2023gqa} share KV states across query heads within a layer, and cross-layer attention~\citep{brandon2405reducing} shares one set of KV states between adjacent layers. Multi-head latent attention (MLA), introduced in DeepSeek-V2 and retained in DeepSeek-V3~\citep{deepseekv2,deepseekv3}, caches one low-rank latent per token per layer with a decoupled RoPE key. Pre-training an MLA model is out of reach for most groups, so TransMLA~\citep{transmla2025} and CARE~\citep{care2026} convert a pretrained GQA checkpoint instead. Orthogonally, KVQuant~\citep{hooper2024kvquant} identifies the per-channel and outlier structure that makes low-bit KV caching viable. Each of these fixes a single structural choice, or one bit-width, at conversion or pre-training time, and fixes it for all layers. Closest to our claim that this kind of compression has to be trained for, \citet{molfese2026exploring} fine-tune long-context models for in-context retrieval and then ask whether that transfers to robustness under KV-cache compression, reporting moderate gains that vary by task. They train for the task and measure compression robustness as a downstream property. We put the budget in the objective and train under the compression itself, across three composed axes.

\paragraph{Cross-layer sharing.} Layers are redundant enough that a full per-layer cache is not obviously necessary~\citep{monteiro2024xc,wu2024layer}. XC-Cache~\citep{monteiro2024xc} shares a cache across layers but needs a separate encoder, so every update costs an encoder pass. Layer-Condensed KV Cache~\citep{wu2024layer} shares a single layer's cache at a cost in time to first token, since approximating the top layer requires repeated forward passes. \textit{Post-hoc} methods avoid retraining at the cost of how far they can go, merging pairs of late layers~\citep{liu2024minicache}, selecting shareable layers by a dissimilarity criterion~\citep{yang2024kvsharer}, sharing projection weights by SVD~\citep{wang2025commonkv}, or training alignment networks for cross-layer differences~\citep{mu2024crosslayer}; \citet{wu2025systematic} evaluate sharing configurations systematically but pretrain each one. xKV~\citep{xkv2026} factorizes the caches of a group of layers into one shared low-rank subspace after training, on the observation that their dominant singular vectors align across layers, which composes our depth and rank axes in a single \textit{post-hoc} step with the grouping and the rank both fixed before any training under them. Closest to our depth axis, \citet{rcla2026} randomize the source layer of the keys and values during fine-tuning so one checkpoint tolerates sharing patterns chosen at test time.

\paragraph{Non-uniform and adaptive allocation.} Several methods accept that a uniform per-layer setting is wasteful. PyramidKV~\citep{cai2024pyramidkv} and CAKE~\citep{qincake} allocate token budgets non-uniformly across layers, and FlexRank~\citep{flexrank2026} decomposes weights into nested low-rank factors so a rank can be chosen per deployment, under an additive cost model solved by a dynamic program. Ada-KV~\citep{adakv2026} and SqueezeAttention~\citep{squeezeattention2025} allocate one eviction budget non-uniformly across heads and across layers. KVTuner~\citep{kvtuner2025} searches offline for a pair of key and value bit-widths per layer under a multi-objective criterion and then serves the configuration it found, which is our precision axis allocated per layer by search rather than learned, and applied to a checkpoint that never trained under it. MatryoshkaKV~\citep{matryoshkakv2025} trains orthogonal projections of the cache features with a distillation objective and a nested schedule, then searches a compression rate per layer and per head for whatever budget it is given, which is our rank axis with a budget that can be set after training. STAR-KV~\citep{starkv2026} selects the rank per head and per block through a differentiable threshold and quantizes the result. Palu~\citep{palu2024} projects the keys and values into a low-rank latent and searches a non-uniform rank per layer against a total compression target, then composes that with quantization. DMC~\citep{dmc2024} retrofits a pretrained model by continued training with a target compression ratio in the loss and a per-token gate trained through a straight-through estimator. Our setting is distinguished by three properties together, none of which is enough on its own. The cache configuration is over a \emph{composed} action space in which depth, rank and precision are chosen jointly. It is made against \emph{one} budget, so configurations are comparable. And the budget can be an input at runtime, so one selector answers a range of requests (\Cref{sec:eval_budget_cond}). To our knowledge, none has placed all these three axes in one action space under one cost model and let a learned selector spend a single budget across them per layer. ThinK~\citep{think2025} prunes key channels per query and composes that with a quantizer, so what it chooses is a function of the prompt where what ours chooses is not. RAT+~\citep{ratplus2026} pretrains one dense model that can be switched at inference to any of several dilated attention patterns, each reducing the cache by its dilation factor.

\section{Experimental details}\label{app:experimental_details}

\paragraph{Backbones and MLA conversion.} \Cref{tab:backbones} lists the backbones and their KV dimensions. For OLMo-3 we use its reasoning variant, \texttt{Olmo-3-7B-Think}. OLMo-3 mixes sliding-window and full attention layers, and a sliding-window layer keeps only its most recent positions, so no other layer can read its cache. We therefore run every OLMo-3 layer with full attention, in every OLMo-3 run including the controls.
 Configurations using rank start from a non-MLA model converted to MLA with CARE~\citep{care2026} and balanced key-value scaling from TransMLA~\citep{transmla2025}. We set $r=n_{\text{kv}}d_h$, $\text{qk\_nope\_head\_dim}=\text{v\_head\_dim}=d_h$, and the RoPE key width $d_R=64$. We retain the \texttt{Qwen2.5} projection biases.

\begin{table}[h]
\centering\small\setlength{\tabcolsep}{5pt}
\caption{\textbf{Backbone dimensions used to compute cache size.} $L$ is the number of layers, $n_{\text{kv}}$ the number of KV heads per layer, and $2n_{\text{kv}}d_h$ the number of cached scalars per token per layer. All models have head dimension $d_h=128$. The converted MLA latent has width $r=n_{\text{kv}}d_h$, excluding the $64$-dimensional positional key.}
\label{tab:backbones}
\begin{tabular}{lrrrr}
\toprule
Backbone & \shortstack{Layers\\$L$} & \shortstack{KV heads\\$n_{\text{kv}}$} & \shortstack{Scalars/token/layer\\$2n_{\text{kv}}d_h$} & \shortstack{MLA width\\$r$} \\
\midrule
\texttt{Qwen2.5-1.5B-Instruct}    & 28 & 2 & 512  & 256  \\
\texttt{Qwen2.5-3B-Instruct}      & 36 & 2 & 512  & 256  \\
\texttt{Qwen2.5-7B-Instruct}      & 28 & 4 & 1024 & 512  \\
\texttt{Qwen2.5-14B-Instruct}     & 48 & 8 & 2048 & 1024 \\
\texttt{Qwen2.5-32B-Instruct}     & 64 & 8 & 2048 & 1024 \\
\texttt{Mistral-7B-Instruct-v0.3} & 32 & 8 & 2048 & 1024 \\
\texttt{Olmo-3-7B-Think}          & 32 & 32 & 8192 & 4096 \\
\bottomrule
\end{tabular}
\end{table}

\paragraph{Quantization.} On non-MLA backbones, we quantize $K_l$ and $V_l$. On MLA backbones, we quantize the retained latent $c_l^{KV}$ and the positional key $k_l^R$ at the same bit-width. We use symmetric uniform quantization with one scale per stored vector, determined by its maximum absolute value. Training quantizes and dequantizes the cache in the forward pass and uses straight-through gradients. \Cref{app:action_space} gives the quantization rule and storage accounting.

\paragraph{Data and optimization.} We fine-tune on Dolci-Instruct-SFT\footnote{\url{https://huggingface.co/datasets/allenai/Dolci-Instruct-SFT}}~\citep{dolci}, applying the chat template, supervising assistant turns only, and truncating sequences to $4096$ tokens. Each run uses $270{,}000$ steps, corresponding to approximately one epoch, with one sequence per data-parallel rank on eight GPUs: eight sequences and at most $32{,}768$ tokens per step. A held-out instruction-following subset provides validation loss and token accuracy.

We optimize the backbone and selector jointly with AdamW~\citep{loshchilov2017decoupled}, using a shared learning rate of $10^{-5}$, linear decay, $2\%$ warmup, weight decay $10^{-4}$, and gradient clipping at $1.0$. Training uses \textsc{bf16} and gradient checkpointing. Models up to $7$B use A100 or H100 nodes; $14$B models use Blackwell nodes, with eight accelerators per job.

\paragraph{Selector architecture.} The selector projects the backbone's token embeddings from width $H$ to $256$ and applies two bidirectional transformer blocks. Each block has four attention heads of width $64$, rotary embeddings on queries and keys, and a SwiGLU feed-forward layer of inner width $1024$. RMSNorm precedes each sublayer, with residual connections around both; a final RMSNorm follows the second block. We average the output over tokens to obtain $z\in\mathbb{R}^{256}$. Each of the $L$ layers has an independent linear head mapping $z$ to $|\mathcal{A}|$ action logits.

With $17$ actions, the selector has $2.61$, $2.78$, $3.14$, $3.62$, and $3.69$M parameters for \texttt{Qwen2.5} at $1.5$, $3$, $7$, $14$, and $32$B, respectively, and $3.29$M for both \texttt{Mistral-7B} and \texttt{OLMo-3-7B}, which share a width and a depth. These are $0.17\%$, $0.09\%$, $0.041\%$, $0.025\%$, and $0.011\%$ of the \texttt{Qwen2.5} backbones, and $0.045\%$ of each of the other two. The per-layer heads account for $122$k--$280$k parameters. \Cref{app:selector_cost} compares alternative architectures.

\paragraph{Budget conditioning.} A FiLM MLP with $33$k parameters maps the normalized target $\tilde{\rho}_\star$ to a scale and shift:
\[
 (\gamma,\beta_{\mathrm F})=\mathrm{MLP}(\tilde{\rho}_\star),
 \qquad z\mapsto z\odot(1+\gamma)+\beta_{\mathrm F}.
\]
Each training sequence samples a target from the training set of compression factors (\Cref{tab:budget-sets}), normalized to approximately $[-1,1]$. We exclude runs whose range of realized compression factors across requested targets is less than $5\%$ of their mean.

\paragraph{Initialization and schedules.} We initialize trunk and head weights from $\mathcal{N}(0,10^{-3})$. Head biases are zero except for a bias of $5.0$ on the least-compressed action. The Gumbel temperature decreases geometrically from $2.0$ to $0.1$ over training, with one sample per sequence at each optimizer step. \Cref{app:budget_controller} specifies the adaptive rate multiplier, whose default cap is $10^4$. For constant-multiplier runs ($\eta=0$), we increase $\beta$ linearly to the selected value during warmup.

\paragraph{Evaluation.} We rely on the lm-evaluation-harness~\citep{eval-harness}. GSM8K scores use the flexible-extract metric. Decoding is greedy on fixed evaluation sets. \Cref{tab:seeds} reports variation across training seeds.

\paragraph{Timing.} We measure pre-fill and selector latency at batch size one on an H100, using precision and depth at a compression factor of $4\times$. We fix the scaled-dot-product attention kernel, synchronize the device before and after each timed region, discard four warmups, and report the median of $12$ repetitions. The reference uses the same weights, kernel, and batch size without the selector. \Cref{app:selector_cost} reports latency and training cost.

\section{Action space and cache storage}\label{app:action_space}

\subsection{Compression ranges}
\Cref{tab:ranges} lists the available actions and compression ranges for two backbone geometries. Disabling depth masks the inherit action. The minimum compression factor uses the highest-cost action at every layer. The maximum uses the lowest-cost action at every layer when depth is disabled, or at a single anchor when depth is enabled:
\begin{equation}\label{eq:range}
  \rho_{\min} = \frac{C_0}{L \max_a c(a)},
  \qquad
  \rho_{\max} =
  \begin{cases}
    C_0/\min_a c(a) & \text{depth enabled},\\
    C_0/(L\min_a c(a)) & \text{depth disabled}.
  \end{cases}
\end{equation}
The extrema are over actions that produce a cache. We check fixed training targets against this range. At the precision-only maximum of $8\times$, every layer must use two bits, leaving only one feasible configuration.

\begin{table}[h]
\centering\small\setlength{\tabcolsep}{5pt}
\caption{\textbf{Action counts and feasible cache reduction factors.} Approaches lists the enabled compression axes; backbone identifies the attention architecture. Actions counts choices per layer, Plans counts assignments across $L$ layers before budget and first-layer constraints, and $[\rho_{\min},\rho_{\max}]$ gives minimum and maximum cache reduction relative to the original \textsc{bf16} GQA cache. The two models have $(L,2n_{\text{kv}}d_h,r)=(48,2048,1024)$ and $(36,512,256)$. Rank uses MLA with positional-key width $d_R=64$. Bit-widths are $\{2,4,8,16\}$ and latent widths $\{r/8,r/4,r/2,r\}$.}
\label{tab:ranges}
\begin{tabular}{llrrrrr}
\toprule
& & & \multicolumn{2}{c}{\bf \texttt{Qwen2.5-14B}} & \multicolumn{2}{c}{\bf \texttt{Qwen2.5-3B}} \\
\cmidrule(r){4-5}\cmidrule(l){6-7}
Approaches & Backbone & Actions & Plans & $[\rho_{\min},\rho_{\max}]$ & Plans & $[\rho_{\min},\rho_{\max}]$ \\
\midrule
Precision                 & GQA & 4  & $10^{29}$ & $1.00$--$8.0\times$  & $10^{22}$ & $1.00$--$8.0\times$ \\
Depth                     & GQA & 2  & $10^{14}$ & $1.00$--$48.0\times$ & $10^{11}$ & $1.00$--$36.0\times$ \\
Depth + precision         & GQA & 5  & $10^{34}$ & $1.00$--$384\times$  & $10^{25}$ & $1.00$--$288\times$ \\
\midrule
Rank                      & MLA & 4  & $10^{29}$ & $1.88$--$10.7\times$ & $10^{22}$ & $1.60$--$5.3\times$ \\
Rank + precision          & MLA & 16 & $10^{58}$ & $1.88$--$85.3\times$ & $10^{43}$ & $1.60$--$42.7\times$ \\
Depth + rank              & MLA & 5  & $10^{34}$ & $1.88$--$512\times$  & $10^{25}$ & $1.60$--$192\times$ \\
Depth + rank + precision  & MLA & 17 & $10^{59}$ & $1.88$--$4096\times$ & $10^{44}$ & $1.60$--$1536\times$ \\
\bottomrule
\end{tabular}
\end{table}

The positional key limits rank compression because truncating the latent does not reduce $d_R$. MLA conversion itself reduces cache size, so its compression factors start above $1\times$. All factors in \Cref{tab:ranges} include this initial reduction.

\subsection{MLA cache representation}\label{app:worked}

An MLA layer stores a latent $c^{KV}$ and a positional key $k^R$ for each token. The content key is reconstructed from the latent using the layer's projection weights. Since it does not undergo RoPE, this projection can be absorbed into the query projection. The positional key is stored after RoPE and shared across heads. The cached width is therefore $d+d_R$, with full widths $512+64$ at $7$B and $1024+64$ at $14$B.

During training, rank truncation uses a prefix mask: coordinates after $d$ are set to zero while the tensor retains its full shape, which allows different widths within a batch. A deployed kernel would instead write only the first $d$ coordinates. We use no explicit nested-latent objective; the representation is initialized by conversion and updated during fine-tuning. Training computes all candidate actions for the straight-through estimator. With gradients disabled, it computes only the selected action; unselected actions have zero weight.

\subsection{Quantization and storage overhead}

Our symmetric quantizer uses one scale per stored vector and no zero-point. For bit-width $b$, let $q=2^{b-1}-1$ and $s=\max_i|x_i|/q$. The integer code is $\operatorname{clip}(\operatorname{round}(x/s),-q,q)$, and dequantization multiplies it by $s$. At two bits, this gives three levels. A grouped asymmetric quantizer instead stores a scale and zero-point per group of $g$ channels and uses all $2^b$ levels.

\begin{table}[h]
\centering\small\setlength{\tabcolsep}{5pt}
\caption{\textbf{Storage cost and cache reduction with quantization metadata.} $b$ is the quantization bit-width; bits/element includes metadata and any full-precision residual window. $\rho$ is $16$ divided by this storage cost, relative to \textsc{bf16}. Our quantizer stores one \textsc{bf16} scale per $128$-element token-head row. KIVI stores a scale and zero-point per group of $g$ elements and a $128$-token \textsc{bf16} window in a $2048$-token context. Bold highlights the factors used in the text.}
\label{tab:realized_bits}
\begin{tabular}{llcc}
\toprule
Scheme & Bit-width $b$ & Bits/element $\downarrow$ & $\rho\uparrow$ \\
\midrule
Our quantizer & 8 & 8.125 & 1.97$\times$ \\
Our quantizer & 4 & 4.125 & \textbf{3.88}$\times$ \\
Our quantizer & 2 & 2.125 & 7.53$\times$ \\
\midrule
KIVI, $g$=32 & 4 & 5.688 & 2.81$\times$ \\
KIVI, $g$=64 & 4 & 5.219 & 3.07$\times$ \\
KIVI, $g$=128 & 4 & 4.984 & 3.21$\times$ \\
KIVI, $g$=32 & 2 & 3.812 & 4.20$\times$ \\
KIVI, $g$=64 & 2 & 3.344 & 4.79$\times$ \\
KIVI, $g$=128 & 2 & 3.109 & \textbf{5.15}$\times$ \\
\bottomrule
\end{tabular}
\end{table}

\Cref{tab:realized_bits} includes quantization metadata and the unquantized residual window. At two bits and $g=128$, KIVI reaches $5.15\times$ compression for a $2048$-token context, or $7.11\times$ without the window. At $16$k, the fixed $128$-token window accounts for a smaller fraction of the cache: four-bit KIVI with $g=128$ uses $4.34$ bits per element, giving the $3.7\times$ factor in \Cref{tab:longctx}.

\section{Setting the rate multiplier $\beta$}\label{app:budget_controller}

\subsection{Update rule}
We adapt the rate multiplier in \cref{eq:rate} to the difference between requested and realized compression. At each training step, we compute $v=\rho_\star/\rho-1$, which is positive when the cache exceeds its target size, and update
\[
 \overline{|v|}\leftarrow\lambda\overline{|v|}+(1-\lambda)|v|,
 \qquad
 \beta\leftarrow\mathrm{clamp}\!\left(\beta+\eta(\overline{|v|}-\delta),0,\beta_{\max}\right).
\]
We use $\lambda=0.9$, $\eta=0.05$, $\delta=0.02$, $\beta_{\max}=10^4$, and initialize $\beta=0$. The multiplier is all-reduced across data-parallel ranks before the update.

\subsection{Budget accuracy and task performance}
The adaptive rule reaches the target in all $51$ evaluated fixed-target configurations. Constant multipliers are less reliable near the maximum feasible compression: with only precision on \texttt{Qwen2.5-7B}, tested constant settings reach $6.32\times$ rather than the requested $8\times$, for instance.

\Cref{tab:swept} compares adaptive and swept constant multipliers at matched realized compression. Six pairs differ by at most $0.053$ in accuracy, the largest observed range across training seeds. The swept constant performs better in the other four, including all three precision-and-depth pairs. Reaching the target budget therefore does not guarantee the best task accuracy. One limitation is slow decay: the update decreases $\beta$ by at most $\eta\delta=0.001$ per step, so a large multiplier can persist after the target has been reached.

We thus cap the multiplier at $\beta_{\max}=50$ to test this, at realized factors within $2\%$ of the target. On \texttt{Qwen2.5-7B} with precision and depth, the cap raises accuracy from $0.255$ to $0.496$ at $8\times$ and from $0.162$ to $0.327$ at $16\times$, and on \texttt{Qwen2.5-14B} at $16\times$ from $0.322$ to $0.535$. With precision and rank, which do not use depth, the same cap raises accuracy by $0.049$ at $8\times$ and $0.055$ at $16\times$. With all three axes, the capped selector inherits on no layer at $8\times$ and on one of $28$ at $16\times$. At $14$B and $16\times$, the cap changes accuracy by $+0.005$ with all three approaches, and almost no layer inherits.

\begin{table}[h]
\centering\small\setlength{\tabcolsep}{5pt}
\caption{\textbf{Mean IFEval and GSM8K accuracy with adaptive and swept constant rate multipliers.} Rows identify the compression axes and \texttt{Qwen2.5} model size; $\rho_\star$ is the target cache reduction factor. Adaptive and Swept give accuracies; the last column subtracts adaptive from swept accuracy. Pairs differ by at most $2\%$ in realized compression. The upper six accuracy differences are within $0.053$, the largest observed seed range.}
\label{tab:swept}
\begin{tabular}{llrrr}
\toprule
& & \multicolumn{2}{c}{Mean accuracy $\uparrow$} & \\
\cmidrule(lr){3-4}
Approaches & Target $\rho_\star$ & Adaptive & Swept & Accuracy difference $\uparrow$ \\
\midrule
All three, $7$B          & $8\times$  & $0.541$ & $0.534$ & $-0.007$ \\
Precision, $7$B          & $4\times$  & $0.670$ & $0.668$ & $-0.002$ \\
All three, $14$B         & $16\times$ & $0.610$ & $0.571$ & $-0.039$ \\
All three, $7$B          & $16\times$ & $0.429$ & $0.448$ & $+0.019$ \\
All three, $14$B         & $8\times$  & $0.622$ & $0.644$ & $+0.022$ \\
All three, $7$B          & $4\times$  & $0.547$ & $0.579$ & $+0.032$ \\
\midrule
All three, $3$B          & $4\times$  & $0.327$ & $0.484$ & $+0.157$ \\
Precision + depth, $7$B  & $16\times$ & $0.162$ & $0.388$ & $+0.226$ \\
Precision + depth, $14$B & $16\times$ & $0.322$ & $0.563$ & $+0.241$ \\
Precision + depth, $7$B  & $8\times$  & $0.255$ & $0.562$ & $+0.307$ \\
\bottomrule
\end{tabular}
\end{table}

\begin{figure}[h]
\begin{center}
\includegraphics[width=\textwidth]{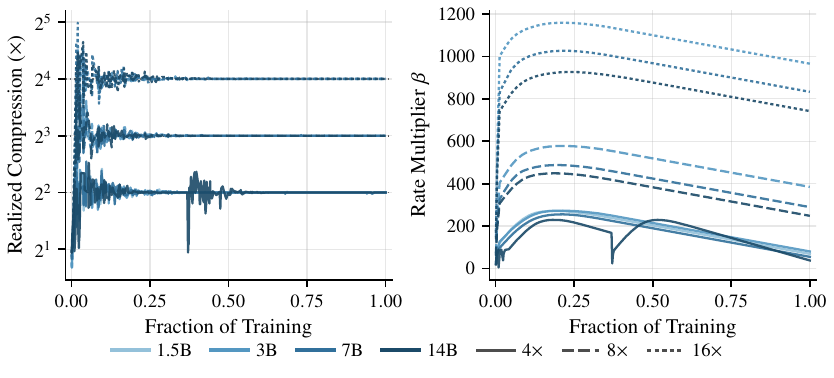}
\end{center}
\caption{\textbf{Compression factor (left) and rate multiplier (right) during training.} Ten runs use all three axes on \texttt{Qwen2.5} at $1.5$, $3$, $7$, and $14$B, with targets $4\times$, $8\times$, and $16\times$; $1.5$B is evaluated only at $4\times$. Color denotes model size and line style denotes target. Training progress is normalized by $270{,}000$ steps.}
\label{fig:training-curves}
\end{figure}

\Cref{fig:training-curves} shows representative runs with all three axes. Compression stabilizes near the target by about $30\%$ of training. The multiplier initially increases and then decreases as the target is reached.

\section{Comparing cache configurations at a fixed budget}\label{app:regret_diagnostic}

\subsection{Alternatives to the learned configuration}
We check whether the selector could have picked a better configuration for the model it trained with. For each trained $7$B model we hold the weights fixed and score $24$ other configurations of about the same cache cost on $128$ held-out prompts. \Cref{tab:regret} answers three questions. \emph{Loss range} asks whether the choice matters at all: it is how far the alternatives' losses spread on a prompt, averaged over prompts. \emph{Per-group loss gain} asks whether choosing a different configuration for each group of prompts would help: the groups' configurations are chosen on half the prompts and scored on the other half. \emph{Mean regret} asks whether any alternative beats the selector: it is the selector's loss minus the best alternative's, so a negative value means none does.

\begin{table}[h]
\centering\small\setlength{\tabcolsep}{5pt}
\caption{\textbf{Loss differences between learned and alternative cache configurations.} We score $24$ alternatives on $128$ prompts with trained $7$B weights fixed and cache costs approximately matched. Negative regret favors the selector. All differences are in nats.}
\label{tab:regret}
\begin{tabular}{lccc}
\toprule
Approaches & Loss range & Per-group loss gain $\uparrow$ & Mean regret $\downarrow$ \\
\midrule
Depth only & $1.4826$ & $+0.0000$ & $-0.4988$ \\
All three  & $0.3014$ & $+0.0048$ & $-0.0143$ \\
\bottomrule
\end{tabular}
\end{table}

No alternative beats the selector in either action space, and choosing per prompt group gains almost nothing. The choice matters most with only depth available, where the alternatives' losses spread by $1.48$ and the selector is $0.50$ below the best of them. With all three approaches the spread is $0.30$ and the margin $0.014$. Because the weights are those trained with the selector, this does not show which configuration would be best if each were trained on its own. \Cref{tab:static} compares that on downstream tasks.

\subsection{Offline search}
We check how much the choice of configuration matters when the cache size is fixed. To do so, we search for a configuration on the fine-tuned, uncompressed MLA control. We compress one layer at a time with each action to estimate how much each costs, and a dynamic program combines these estimates into the configuration with the lowest estimated loss that fits the budget. We compare it with random configurations and with a uniform one, which applies the same action to every layer.

\begin{table}[h]
\centering\small\setlength{\tabcolsep}{5pt}
\caption{\textbf{Loss increase from compressing the uncompressed MLA control of \texttt{Qwen2.5-7B} without training.} Measured is the rise in teacher-forced loss on calibration data; predicted is the sum of per-layer rises that the search minimizes. Random is the best of $64$ random configurations at the same budget. No uniform configuration meets the $4\times$ budget exactly, and dashes mark values not measured. All values are in nats.}
\label{tab:offline}
\begin{tabular}{lccc}
\toprule
& \multicolumn{2}{c}{Measured $\downarrow$} & Predicted \\
\cmidrule(lr){2-3}\cmidrule(lr){4-4}
Configuration & $4\times$ & $8\times$ & $8\times$ \\
\midrule
Searched & $0.005$ & $0.030$ & $0.022$ \\
Random   & $0.125$ & $1.253$ & -- \\
Uniform  & --      & $4.00$  & $0.264$ \\
\bottomrule
\end{tabular}
\end{table}

At the same cache size, the searched configuration barely changes the loss, random ones raise it clearly, and the uniform one raises it far more (\Cref{tab:offline}). The per-layer estimates predict the searched configuration well but miss most of the uniform one's loss, because when every layer is compressed too much, the damage no longer adds up layer by layer. Which configuration is used therefore matters far more than the cache size alone, which is why we choose it per layer.

\section{Additional experimental results}\label{app:full_results}

This appendix collects the results that support \Cref{sec:eval}. Each subsection states what it checks and what it finds, and gives the full tables behind numbers the main text quotes.

\subsection{Quantization and smaller-model baselines}
We check whether the same cache reduction can be had without training, either by quantizing a model after training or by serving a smaller model. \Cref{tab:posthoc} compares the three against the uncompressed $7$B control, with every cache size relative to that model. Applying our plain quantizer after training destroys the model, while training under the same quantizer keeps it level with the control, so the quantizer is not the problem; the model has to train under it. A stronger post-hoc quantizer, KIVI, holds up at four bits but starts to lose accuracy at two bits. Serving a smaller uncompressed model is the weakest option: \texttt{Qwen2.5-1.5B} saves only $2\times$ and falls far below a $7$B model compressed $4\times$.

\begin{table}[h]
\centering\small\setlength{\tabcolsep}{5pt}
\caption{\textbf{Cache reduction and task accuracy with quantization, trained compression, and smaller models.} $\rho$ is the cache reduction factor relative to the full \texttt{Qwen2.5-7B} \textsc{bf16} cache. IFEval and GSM8K report task accuracies on $[0,1]$; Mean is their arithmetic mean. Parentheses give mean-accuracy differences from the SFT control; bold marks means within $0.053$ of that control or higher. $G$ is the quantization group size. Post-hoc factors include metadata; trained precision gives $4.00\times$ before metadata and $3.88\times$ after it.}
\label{tab:posthoc}
\begin{tabular}{llccc}
\toprule
& & \multicolumn{3}{c}{Accuracy $\uparrow$} \\
\cmidrule(lr){3-5}
Configuration & $\rho\uparrow$ & IFEval & GSM8K & Mean \\
\midrule
\texttt{Qwen2.5-7B-Instruct}, base & 1.00$\times$ & 0.706 & 0.701 & 0.704 \\
\midrule
Our SFT, uncompressed (control) & 1.00$\times$ & 0.719 & 0.639 & 0.679 \\
\quad + post-hoc 4-bit, plain & 3.88$\times$ & 0.113 & 0.001 & 0.057\,($-0.622$) \\
\quad + \emph{trained under} KIVI, 4-bit $G$=32 & 2.81$\times$ & 0.701 & 0.680 & \textbf{0.690}\,($+0.011$) \\
\quad + post-hoc KIVI, 4-bit $G$=32 & 2.81$\times$ & 0.721 & 0.635 & \textbf{0.678}\,($-0.001$) \\
\quad + post-hoc KIVI, 4-bit $G$=128 & 3.21$\times$ & 0.723 & 0.638 & \textbf{0.680}\,($+0.001$) \\
\quad + post-hoc KIVI, 2-bit $G$=128 & 5.15$\times$ & 0.713 & 0.590 & \textbf{0.652}\,($-0.027$) \\
\quad + trained precision & 4.00$\times$ & 0.723 & 0.617 & \textbf{0.670}\,($-0.009$) \\
\quad + trained precision + depth & 4.00$\times$ & 0.721 & 0.637 & \textbf{0.679}\,($-0.000$) \\
\midrule
\texttt{Qwen2.5-3B}, uncompressed & 1.56$\times$ & 0.706 & 0.641 & 0.673 \\
\texttt{Qwen2.5-1.5B}, uncompressed & 2.00$\times$ & 0.645 & 0.509 & 0.577 \\
\bottomrule
\end{tabular}
\end{table}

\FloatBarrier
\subsection{Accuracy by compression axis}
\Cref{tab:axes-7b} breaks the $7$B column of \Cref{tab:axes} down per task, to show where each approach loses accuracy. Converting to MLA costs accuracy, so configurations using rank are read against the converted control, which separates the choice from the conversion. Against that reference, every configuration that uses rank or precision stays within the seed range. Depth on its own is the exception, and it loses far more on reasoning than on instruction following.  

\begin{table}[h]
\centering\small\setlength{\tabcolsep}{5pt}
\caption{\textbf{Task accuracy by compression axis on \texttt{Qwen2.5-7B} at $4\times$ cache reduction.} IFEval and GSM8K report task accuracies on $[0,1]$; Mean is their arithmetic mean. Ref. identifies the control used for parenthesized mean-accuracy differences: MLA for configurations using rank, GQA otherwise. The conversion-only row is compared with GQA. Bold marks means within $0.053$ of the corresponding control or higher}
\label{tab:axes-7b}
\begin{tabular}{llccc}
\toprule
& & \multicolumn{3}{c}{Accuracy $\uparrow$} \\
\cmidrule(lr){3-5}
Approaches & Ref. & IFEval & GSM8K & Mean \\
\midrule
Uncompressed (GQA) & -- & 0.719 & 0.639 & 0.679 \\
\quad MLA conversion only & GQA & 0.636 & 0.513 & 0.574\,($-0.105$) \\
\midrule
Precision & GQA & 0.723 & 0.617 & \textbf{0.670\,($-0.009$)} \\
Depth & GQA & 0.547 & 0.245 & 0.396\,($-0.283$) \\
Precision + depth & GQA & 0.721 & 0.637 & \textbf{0.679\,($-0.000$)} \\
Rank & MLA & 0.601 & 0.492 & \textbf{0.546\,($-0.028$)} \\
Precision + rank & MLA & 0.619 & 0.522 & \textbf{0.571\,($-0.003$)} \\
Depth + rank & MLA & 0.593 & 0.473 & \textbf{0.533\,($-0.041$)} \\
All three & MLA & 0.632 & 0.461 & \textbf{0.547\,($-0.028$)} \\
\bottomrule
\end{tabular}
\end{table}

\FloatBarrier
\subsection{Long-context evaluation}

\paragraph{Eviction implementations.} Our SnapKV~\citep{li2024snapkv}, PyramidKV~\citep{cai2024pyramidkv}, and H\textsubscript{2}O\citep{zhang2023h2o} implementations sum attention over a $32$-query observation window. Published H\textsubscript{2}O accumulates attention over all past queries; our pre-fill implementation is therefore a restricted variant. StreamingLLM~\citep{xiao2024efficient} retains attention sinks and a recent-token window. Recency and random selection do not use attention scores.

\paragraph{Results at $\bm{7}$B.} We repeat the long-context comparison at $7$B, with four further eviction policies (\Cref{tab:longctx-7b}). Our $4\times$ configuration is at the same level as query-informed SnapKV and PyramidKV, and combined with SnapKV at $16\times$ it is at the same level as four-bit KIVI under the same evictor. The policies that choose which tokens to keep before the query arrives fall far behind, as does using depth or rank on its own at this context length.

\begin{table}[h]
\centering\small\setlength{\tabcolsep}{5pt}
\caption{\textbf{RULER accuracy at $\bm{16}$k on \texttt{Qwen2.5-7B}.} $\rho$ is the decode-time cache reduction factor relative to the uncompressed GQA model. NIAH single and multi-key report retrieval accuracy; QA reports SQuAD question-answering accuracy. Parentheses give QA differences from the control. Keep is the retained token fraction, and $G$ the quantization group size. Rank uses MLA; other rows use GQA. Bold identifies \kvcubeshort{} configurations and their combinations with eviction. The last four policies select tokens before the query arrives.}
\label{tab:longctx-7b}
\begin{tabular}{lrccc}
\toprule
& & \multicolumn{3}{c}{Accuracy $\uparrow$} \\
\cmidrule(lr){3-5}
Configuration & $\rho\uparrow$ & NIAH single & NIAH multi-key & QA (SQuAD) \\
\midrule
No compression (control) & $1.0$ & 1.000 & 0.985 & 0.540 \\
\midrule
\textbf{\kvcubeshort}, $4\times$ (precision + depth) & $4.0$ & 0.996 & 0.956 & 0.510\,($-0.030$) \\
\quad \textbf{+ SnapKV}, keep $0.25$ & $16.0$ & 0.985 & 0.887 & 0.472\,($-0.068$) \\
\textbf{\kvcubeshort}, $3.9\times$ (precision only) & $3.9$ & 0.978 & 0.913 & 0.508\,($-0.032$) \\
\textbf{\kvcubeshort}, $4\times$ (depth only) & $4.0$ & 0.964 & 0.324 & 0.107\,($-0.434$) \\
\textbf{\kvcubeshort}, $4\times$ (rank only, converted) & $4.0$ & 0.978 & 0.247 & 0.165\,($-0.376$) \\
\midrule
KIVI 4-bit $G$=32 & $3.1$ & 1.000 & 0.989 & 0.545\,($+0.005$) \\
KIVI 4-bit $G$=128 & $3.7$ & 1.000 & 0.978 & 0.534\,($-0.006$) \\
KIVI 4-bit $G$=128 + SnapKV $0.25$ & $14.8$ & 1.000 & 0.865 & 0.474\,($-0.067$) \\
\midrule
SnapKV, keep $0.25$ & $4.0$ & 1.000 & 0.891 & 0.474\,($-0.066$) \\
PyramidKV, keep $0.25$ & $4.0$ & 1.000 & 0.807 & 0.469\,($-0.072$) \\
StreamingLLM, keep $0.25$ & $4.0$ & 0.229 & 0.305 & 0.272\,($-0.268$) \\
Recency, keep $0.25$ & $4.0$ & 0.229 & 0.305 & 0.242\,($-0.299$) \\
H\textsubscript{2}O, keep $0.25$ & $4.0$ & 0.331 & 0.000 & 0.396\,($-0.144$) \\
Random, keep $0.25$ & $4.0$ & 0.000 & 0.000 & 0.179\,($-0.362$) \\
\bottomrule
\end{tabular}
\end{table}

\paragraph{Splitting a larger reduction at $\bm{14}$B.} We ask how to spend a large decode-time cache reduction at $16$k: on our compression, on eviction, or on both. Splitting it works best (\Cref{tab:longctx-split}). At $32\times$, a $4\times$ configuration with eviction keeping one position in eight stays level with the uncompressed control, an $8\times$ configuration with eviction keeping one in four falls behind it, and reaching $16\times$ with our compression alone collapses question answering. Under the same evictor, our configuration also beats four-bit KIVI at every retained fraction, each time with a smaller cache. This panel uses the non-MLA backbone because converting to MLA alone costs most of the question-answering accuracy at this context length.

\begin{table}[h]
\centering\small\setlength{\tabcolsep}{5pt}
\caption{\textbf{Question answering at $16$k on \texttt{Qwen2.5-14B} when a larger cache reduction is split between our compression and eviction.} $\rho$ is the decode-time cache reduction relative to the uncompressed GQA model, and keep the fraction of positions SnapKV retains. \kvcubeshort{} configurations use precision and depth. Parentheses give differences from the uncompressed control; the $32\times$ configuration with keep $0.125$ is $0.012$ below it. In the lower block, \kvcubeshort{} and four-bit KIVI with $G=128$ share the evictor, and \kvcubeshort{} is ahead by $0.098$, $0.074$ and $0.067$.}
\label{tab:longctx-split}
\begin{tabular}{lcc}
\toprule
Configuration & $\rho\uparrow$ & QA (SQuAD) $\uparrow$ \\
\midrule
Uncompressed control & $1.0$ & $0.543$ \\
\midrule
\multicolumn{3}{l}{\emph{$32\times$ in total}} \\
\kvcubeshort{} $4\times$ + SnapKV, keep $0.125$ & $32.0$ & $0.531\,(-0.012)$ \\
\kvcubeshort{} $8\times$ + SnapKV, keep $0.25$  & $32.0$ & $0.491\,(-0.052)$ \\
\midrule
\multicolumn{3}{l}{\emph{Without eviction}} \\
\kvcubeshort{} $16\times$ & $16.0$ & $0.258\,(-0.285)$ \\
MLA conversion only       & $1.9$  & $0.210\,(-0.333)$ \\
\midrule
\multicolumn{3}{l}{\emph{\kvcubeshort{} $4\times$ / four-bit KIVI, same evictor}} \\
Keep $0.5$   & $8.0$ / $7.4$   & $0.595$ / $0.498$ \\
Keep $0.25$  & $16.0$ / $14.8$ & $0.560$ / $0.486$ \\
Keep $0.125$ & $32.0$ / $29.5$ & $0.531$ / $0.464$ \\
\bottomrule
\end{tabular}
\end{table}

\FloatBarrier
\subsection{Effect of fine-tuning}\label{app:sft_vs_base}
\begin{table}[h]
\centering\small\setlength{\tabcolsep}{5pt}
\caption{\textbf{Mean IFEval and GSM8K accuracy before and after SFT without cache compression.} Untuned and After our SFT report arithmetic means of the two task accuracies. The Accuracy change is the score after SFT minus the untuned score, computed before rounding.}
\label{tab:sft}
\begin{tabular}{lrrr}
\toprule
& \multicolumn{2}{c}{Mean accuracy $\uparrow$} & \\
\cmidrule(lr){2-3}
Backbone & Untuned & After our SFT & Accuracy change \\
\midrule
\texttt{Qwen2.5-0.5B} & 0.297 & 0.436 & $+0.139$ \\
\texttt{Qwen2.5-1.5B} & 0.488 & 0.577 & $+0.090$ \\
\texttt{Qwen2.5-3B}   & 0.581 & 0.673 & $+0.093$ \\
\texttt{Qwen2.5-7B}   & 0.704 & 0.679 & $-0.025$ \\
\texttt{Qwen2.5-14B}  & 0.774 & 0.723 & $-0.052$ \\
\texttt{Qwen2.5-32B}  & 0.779 & 0.793 & $+0.014$ \\
\texttt{Mistral-7B}   & 0.501 & 0.349 & $-0.152$ \\
\bottomrule
\end{tabular}
\end{table}

SFT improves the smaller Qwen models, while changes at $7$B and above range from $-0.052$ to $+0.014$. Mistral loses $0.152$. These differences motivate reporting compression relative to a control trained with the same recipe.

\FloatBarrier
\subsection{Learned and static cache configurations}

Here, we test two parts of the per-layer choice separately (\Cref{tab:depth-vs-written}). The first is where to place the layers that keep their own cache once their number is fixed: with only depth available at $4\times$, three layers in four must inherit, and we compare the learned placement with evenly spaced and randomly drawn anchors. The second is whether layers should share at all: with all three approaches available, we compare the learned configuration with one that forces every second layer to inherit and one that allows no sharing.

The three placements are close to each other at $3$B, and evenly spaced anchors do better than learned ones at $7$B. What costs seems to be how many layers inherit, not which. Simple sharing patterns can therefore place anchors when sharing is required, but whether to share is worth learning, since precision and rank can often meet the budget without it, and a learned selector can rely on depth only if needed.

\begin{table}[h]
\centering\small\setlength{\tabcolsep}{5pt}
\caption{\textbf{Task accuracy with learned and fixed cross-layer sharing.} Inheriting fraction is the fraction of layers reading another layer's cache; $4\times$ and $8\times$ columns give arithmetic mean IFEval and GSM8K accuracy at those cache reduction targets. Control rows give uncompressed accuracy. Parentheses subtract the block's control; dashes indicate unavailable comparisons. Bold marks scores within $0.053$ of the best compressed configuration in each block and column, excluding the starred entry. In the third block, learned, fixed-sharing, and no-sharing runs realize $4.00$, $3.94$, and $4.00\times$, respectively, and $8.00\times$ at the higher target. $^*$Constant rate multiplier; other compressed rows use the adaptive rule.}
\label{tab:static}
\label{tab:forced-depth}
\label{tab:depth-vs-written}
\begin{tabular}{llcc}
\toprule
& & \multicolumn{2}{c}{Mean accuracy $\uparrow$ at target $\rho_\star$} \\
\cmidrule(lr){3-4}
Sharing configuration & Inheriting fraction & $4\times$ & $8\times$ \\
\midrule
\multicolumn{4}{l}{\emph{Which layers hold the anchors: only depth available, GQA backbone, \texttt{Qwen2.5-3B}}} \\
Uncompressed control          & 0.000 & 0.673 & -- \\
Learned                       & 0.750 & \textbf{0.320\,($-0.354$)} & -- \\
Spaced evenly                 & 0.750 & \textbf{0.339\,($-0.334$)} & -- \\
Drawn at random               & 0.750 & \textbf{0.360\,($-0.313$)} & -- \\
\midrule
\multicolumn{4}{l}{\emph{Which layers hold the anchors: only depth available, GQA backbone, \texttt{Qwen2.5-7B}}} \\
Uncompressed control          & 0.000 & 0.679 & -- \\
Learned                       & 0.750 & 0.396\,($-0.283$) & -- \\
Spaced evenly                 & 0.750 & \textbf{0.452\,($-0.227$)} & -- \\
Drawn at random               & 0.750 & \textbf{0.437\,($-0.242$)} & -- \\
\midrule
\multicolumn{4}{l}{\emph{How many layers inherit: all three axes, MLA backbone, \texttt{Qwen2.5-7B}}} \\
Conversion only (control)     & 0.000 & 0.574 & 0.574 \\
Learned                       & 0.000 & $\mathbf{0.547}$ & $\mathbf{0.541}$ \\
Fixed sharing, every second layer & 0.500 & 0.460 & 0.466 \\
No sharing        & 0.000 & $\mathbf{0.571}$ & 0.558$^{*}$ \\
\bottomrule
\end{tabular}
\end{table}

\FloatBarrier
\subsection{Training seeds and model size}\label{app:seeds_scaling}

\Cref{tab:seeds} reports repeated training runs under different random seeds, to show how much a result moves when only the seed changes. The largest accuracy range, $0.053$, belongs to the uncompressed $7$B control and serves as a descriptive reference.

\begin{table}[h]
\centering\small\setlength{\tabcolsep}{5pt}
\caption{\textbf{Mean IFEval and GSM8K accuracy across training seeds.} Rows identify the \texttt{Qwen2.5} model size and compression axes. $\rho_\star$ is the target cache reduction factor; Seeds counts training runs. Mean accuracy lists the arithmetic mean accuracy for each seed. Bold marks the largest accuracy range, $0.053$. Other tables use the first listed seed.}
\label{tab:seeds}
\begin{tabular}{llcc}
\toprule
Configuration & Target $\rho_\star$ & Seeds & Mean accuracy $\uparrow$ \\
\midrule
Uncompressed, $7$B & $1\times$ & 3 & \textbf{0.679, 0.635, 0.626} \\
Uncompressed, $14$B & $1\times$ & 2 & 0.723, 0.681 \\
\midrule
Precision + depth, $7$B & $4\times$ & 3 & 0.679, 0.679, 0.652 \\
Precision + depth, $7$B & $8\times$ & 3 & 0.255, 0.262, 0.288 \\
Precision + depth, $14$B & $16\times$ & 2 & 0.322, 0.295 \\
All three, $3$B & $4\times$ & 3 & 0.327, 0.334, 0.347 \\
\bottomrule
\end{tabular}
\end{table}

\Cref{tab:scaling} extends the compression comparison to $32$B. The accuracy cost of MLA conversion is not monotonic with model size: accuracy decreases by $0.178$, $0.105$, $0.091$, and $0.131$ at $3$, $7$, $14$, and $32$B. At $32$B, language modeling loss nevertheless improves from $0.448$ to $0.353$, showing that lower loss need not imply higher task accuracy.

\begin{table}[h]
\centering\small\setlength{\tabcolsep}{5pt}
\caption{\textbf{Mean IFEval and GSM8K accuracy differences across \texttt{Qwen2.5} model sizes.} Columns give parameter counts; blocks give realized cache reduction factors. The first two rows report absolute arithmetic mean accuracies after and before SFT. Other entries subtract the size-matched MLA control for configurations using rank, or GQA control otherwise. Bold marks differences $\geq-0.053$; dashes indicate unavailable runs.}
\label{tab:scaling}
\begin{tabular}{lccccc}
\toprule
& \multicolumn{5}{c}{Mean accuracy / difference $\uparrow$} \\
\cmidrule(lr){2-6}
Approaches & 1.5B & 3B & 7B & 14B & 32B \\
\midrule
Uncompressed control & 0.577 & 0.673 & 0.679 & 0.723 & 0.793 \\
Untuned model & 0.488 & 0.581 & 0.704 & 0.774 & 0.779 \\
\midrule
\multicolumn{6}{l}{\emph{At a matched realized $4\times$}} \\
Precision & $-0.034$ & $-0.041$ & $\mathbf{-0.009}$ & $\mathbf{+0.040}$ & $\mathbf{+0.016}$ \\
Precision + depth & -- & $-0.063$ & $\mathbf{-0.000}$ & $\mathbf{+0.031}$ & $\mathbf{+0.030}$ \\
All three & $-0.174$ & $-0.169$ & $\mathbf{-0.028}$ & $\mathbf{+0.021}$ & -- \\
\midrule
\multicolumn{6}{l}{\emph{At a matched realized $8\times$}} \\
All three & -- & $-0.161$ & $\mathbf{-0.033}$ & $\mathbf{-0.010}$ & -- \\
\bottomrule
\end{tabular}
\end{table}

\FloatBarrier
\subsection{Discrete and continuous budget conditioning}\label{app:budget_sets}

\begin{table}[h]
\centering\small\setlength{\tabcolsep}{2.5pt}
\caption{\textbf{Mean IFEval and GSM8K accuracy with selectors trained on wider sets of targets, on \texttt{Qwen2.5-7B}.} $\rho_\star$ is the requested cache reduction factor, and blocks follow \Cref{tab:budget-sets}. Parentheses give the realized factor where a selector misses the request by more than $2\%$. Final $\beta$ is the rate multiplier at the end of training. Dashes mark targets outside the training set or values not recorded.}
\label{tab:budget-width}
\begin{tabular}{lccccccc}
\toprule
& \multicolumn{6}{c}{Mean accuracy $\uparrow$ at requested $\rho_\star$} & \\
\cmidrule(lr){2-7}
Training targets & $1\times$ & $2\times$ & $4\times$ & $8\times$ & $16\times$ & $32\times$ & Final $\beta$ \\
\midrule
\multicolumn{8}{l}{\emph{Precision and depth}} \\
$\{1,2,4\}$          & $0.703$ & $0.659$ & $0.612$ & -- & -- & -- & $355$ \\
$\{1,2,4,8\}$        & $0.672$ & $0.395\,(2.31\times)$ & $0.355$ & $0.353$ & -- & -- & $886$ \\
$\{1,2,4,8,16\}$     & $0.655$ & $0.471\,(2.33\times)$ & $0.269\,(5.47\times)$ & $0.206\,(9.00\times)$ & $0.213$ & -- & $2251$ \\
$\{1,2,4,8,16,32\}$  & $0.585$ & $0.263$ & $0.100\,(6.78\times)$ & $0.107\,(20.4\times)$ & $0.106\,(32.0\times)$ & $0.106$ & $3722$ \\
\midrule
\multicolumn{8}{l}{\emph{All three approaches}} \\
$\{2,4,8\}$          & -- & $0.481$ & $0.463$ & $0.460$ & -- & -- & -- \\
$\{4,8,16\}$         & -- & -- & $0.340\,(3.80\times)$ & $0.319$ & $0.306$ & -- & -- \\
\bottomrule
\end{tabular}
\end{table}

We first check whether it matters if the targets are given as a set of factors or as a continuous range. With all three approaches, a selector trained on the set $\{2,4,8\}$ and one trained over the range $[2,8]$ are within close performance of each other wherever both meet the request (\Cref{tab:budget-width,tab:budget-sets}), so the form of the targets matters little.

We then check whether more training closes the gap to single-target selectors. Doubling the steps helps the range selector at every requested factor, and it reaches near the single-target selectors it replaces, for two thirds of their total training cost in terms of steps (\Cref{tab:budget-sets}).

We finally check why wider sets fail. With depth available, every target in \Cref{tab:budget-width} can be reached, yet the wider sets stop following the request, mostly by compressing more than asked, and their rate multiplier does not settle, ending training up to ten times higher than for $\{1,2,4\}$.

\FloatBarrier
\subsection{Use of cross-layer sharing}

\begin{table}[h]
\centering\small\setlength{\tabcolsep}{5pt}
\caption{\textbf{Fraction of layers that inherit another layer's cache in the selector's configuration.} Ranges span the evaluated model sizes, and the all-three row covers eleven fixed-target configurations. Past $8\times$ on a non-MLA backbone, seven of $73$ trained configurations reach the factor, all with $50\%$ to $94\%$ of layers inheriting. One budget-conditioned configuration with all three approaches inherits at $3.66\times$, for $21.5\%$ of its memory saving. Dashes mark values not recorded.}
\label{tab:sharing}
\begin{tabular}{llccc}
\toprule
& & \multicolumn{3}{c}{Inheriting fraction at target $\rho_\star$} \\
\cmidrule(lr){3-5}
Approaches & Backbone & $4\times$ & $8\times$ & $16\times$ \\
\midrule
Precision + depth & Non-MLA & $0.000$ & $0.021$--$0.107$ & $0.500$--$0.528$ \\
Depth + rank      & MLA     & --      & --              & $0.333$--$0.667$ \\
All three         & MLA     & $0.000$ & $0.000$         & $0.000$ \\
\bottomrule
\end{tabular}
\end{table}

This appendix assesses when the selector uses depth, the approach that lets a layer read another layer's cache (\Cref{tab:sharing}). The answer depends on what else is available. On a non-MLA backbone, precision cannot pass $8\times$, so depth is the only way further, and every configuration we train past that factor lets at least half of its layers inherit. Below it, the selector uses depth little or not at all, and more of it as the budget tightens. With precision and rank both available, it does not use depth at any budget we train with a fixed target, because the two meet those budgets together.

The selector therefore treats depth as the approach of last resort, used when the others cannot meet the budget. This matches \Cref{tab:depth-vs-written}, where imposing sharing that is not needed costs accuracy.

\FloatBarrier
\subsection{Cache configuration diversity}\label{app:replay}

\Cref{tab:prompt-dependence} counts the unique per-layer configurations each selector returns on $64$ IFEval prompts, $64$ GSM8K prompts, and $96$ RULER prompts, $32$ each at $4$k, $8$k, and $16$k tokens. Across model sizes, six selectors return several configurations and six return a single one, re-used across inputs.

\begin{table}[h]
\centering\small\setlength{\tabcolsep}{5pt}
\caption{\textbf{Number of distinct per-layer cache configurations returned by selectors on \texttt{Qwen2.5}.} Rows give the model size and the target cache reduction factor $\rho_\star$. Each entry is the number of distinct configurations a selector returns over the prompts sampled from a dataset, out of the number of prompts: $64$ from IFEval, $64$ from GSM8K, and $96$ from RULER, $32$ at each of $4$k, $8$k, and $16$k tokens. The last column pools all $224$ prompts. Selectors in the upper block use all three approaches. $^\dagger$Rate multiplier capped at $50$. $^\ddagger$Constant rate multiplier; this selector realizes $7.73\times$.}
\label{tab:prompt-dependence}
\begin{tabular}{llcccc}
\toprule
& & \multicolumn{4}{c}{Distinct configurations / prompts} \\
\cmidrule(lr){3-6}
Model size & Target $\rho_\star$ & IFEval & GSM8K & RULER & All three \\
\midrule
\multicolumn{6}{l}{\emph{One configuration}} \\
$3$B  & $4\times$  & $1/64$ & $1/64$ & $1/96$ & $1/224$ \\
$3$B  & $16\times$ & $1/64$ & $1/64$ & $1/96$ & $1/224$ \\
$7$B  & $4\times$  & $1/64$ & $1/64$ & $1/96$ & $1/224$ \\
$7$B  & $8\times$  & $1/64$ & $1/64$ & $1/96$ & $1/224$ \\
$14$B & $4\times$  & $1/64$ & $1/64$ & $1/96$ & $1/224$ \\
$14$B & $16\times$ & $1/64$ & $1/64$ & $1/96$ & $1/224$ \\
\midrule
\multicolumn{6}{l}{\emph{Several configurations}} \\
$3$B  & $2\times$  & $4/64$ & $1/64$ & $6/96$ & $6/224$ \\
$3$B$^\dagger$  & $4\times$  & $2/64$ & $3/64$ & $1/96$ & $3/224$ \\
$7$B$^\dagger$  & $8\times$  & $5/64$ & $2/64$ & $1/96$ & $5/224$ \\
$7$B$^\dagger$  & $16\times$ & $3/64$ & $1/64$ & $1/96$ & $3/224$ \\
$14$B$^\ddagger$ & $8\times$  & $9/64$ & $1/64$ & $13/96$ & $20/224$ \\
$14$B$^\dagger$ & $16\times$ & $4/64$ & $2/64$ & $1/96$ & $4/224$ \\
\bottomrule
\end{tabular}
\end{table}

\FloatBarrier
\subsection{Selector architecture}\label{app:selector_cost}

\paragraph{Architecture comparison.} To check whether the selector's design matters, we compare our transformer over the token embeddings with a two-layer MLP over mean-pooled embeddings and with linear heads over the backbone's last-layer hidden states (\Cref{tab:selector-arch}). With precision only, the three reach similar accuracies. With all three approaches, the hidden-state variant scores higher, at a slightly larger cache, and the MLP falls behind. We keep the transformer because it reads only the prompt's embeddings and so runs before pre-fill, whereas the hidden-state variant needs a full backbone pass to choose the configuration and then a second pre-fill under it.

\begin{table}[h]
\centering\small\setlength{\tabcolsep}{5pt}
\caption{\textbf{Mean IFEval and GSM8K accuracy with three selector architectures on \texttt{Qwen2.5-7B} at $4\times$.} Every selector realizes $4.00\times$ except the hidden-state variant with all three approaches, which realizes $3.86\times$.}
\label{tab:selector-arch}
\begin{tabular}{llcc}
\toprule
& & \multicolumn{2}{c}{Mean accuracy $\uparrow$} \\
\cmidrule(lr){3-4}
Selector & Input & Precision & All three \\
\midrule
Transformer (ours) & Token embeddings & $0.670$ & $0.547$ \\
Two-layer MLP & Mean-pooled embeddings & $0.660$ & $0.510$ \\
Linear heads & Last-layer hidden states & $0.674$ & $0.601$ \\
\bottomrule
\end{tabular}
\end{table}

\paragraph{Training time.} Across $3$, $7$, and $14$B models, training takes $1.0$--$1.4\times$ as long as ordinary fine-tuning with one cache-producing action per layer, $1.3$--$1.8\times$ with four, and $1.9$--$3.1\times$ with sixteen. Each run uses $270{,}000$ steps on eight NVIDIA H100 GPUs. For precision-only \texttt{Qwen2.5-7B}, three separate target-specific runs total $136$ hours, compared with $34.6$ hours for one selector trained on four targets.

\section{On Device benchmarks}\label{app:on_device}

All benchmarks are realized on a M3 Ultra with 512GB of RAM. GPU and CPU temperatures and frequencies are monitored to ensure no thermal throttling take place during measurements.  

\subsection{The Role of Precision}

\begin{figure}
    \centering
    \begin{minipage}[t]{0.48\linewidth}
        \includegraphics[width=0.95\linewidth]{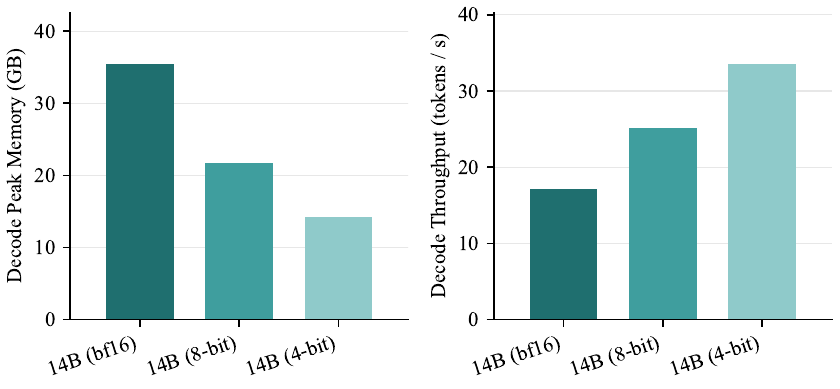}
        \caption{Qwen2.5-14B at three weight precisions, at 27,611-token prompt followed by 2,048 generated tokens. The KV cache is uncompressed bf16 in all three.}
        \label{fig:modelprecision}
    \end{minipage}
    \hfill
    \begin{minipage}[t]{0.48\linewidth}
        \includegraphics[width=0.95\linewidth]{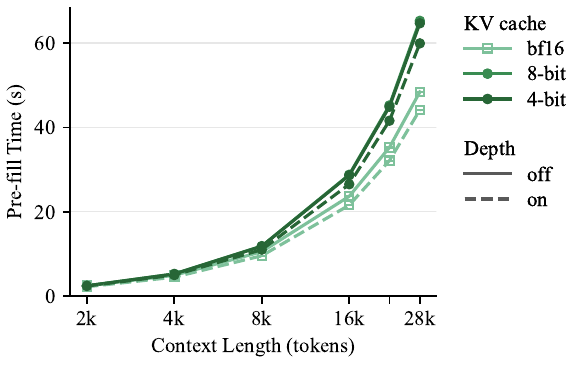}
        \caption{Qwen2.5-14B (4 bits weights) at three KV-cache precision.}
        \label{fig:kvprecision}
    \end{minipage}
\end{figure}

\paragraph{KV-cache Precision.} In MLX~\citep{mlx2023} (version 0.32.2, September 2026), the Scaled Dot-Product Attention (SDPA) relies on a Flash kernel for bf16 weights, but for quantized matrices SDPA is decomposed into a set of quantized matmul that do not benefit from Flash: this increases the cost of Self Attention in low-precision for long sequences as illustrated in \autoref{fig:kvprecision}.  
  
\paragraph{Model Weight Precision.} As shown in~\autoref{fig:modelprecision}, for fixed KV-cache precision and a fixed sequence length (around 30K), the weight precision conserve a massive impact on the total throughput and total memory footprint. This axis is orthogonal to KV-cache precision.  

\begin{figure}
    \centering
    \includegraphics[width=0.5\linewidth]{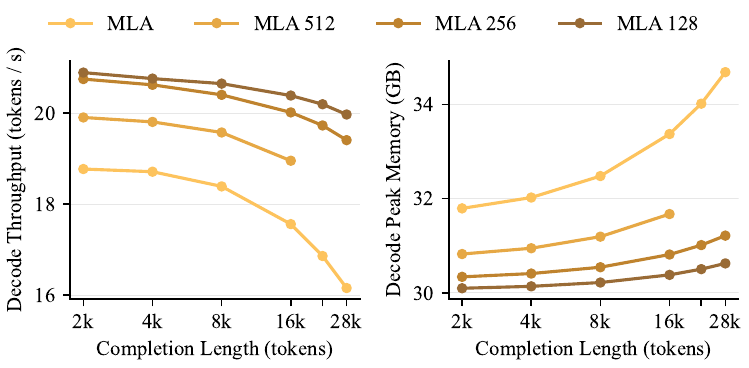}
    \caption{Qwen2.5-14B (4 bits weights) with bf16 KV-cache.}
    \label{fig:mlarank}
\end{figure}

\paragraph{MLA rank.} As shown in~\autoref{fig:mlarank} the Decode Throughput increases when MLA rank is reduced, as one would expect.  

\end{document}